\documentclass{article}

\usepackage{microtype}
\usepackage{graphicx}
\usepackage{booktabs} 
\usepackage{multirow}
\usepackage{graphicx}
\usepackage{enumitem}
\usepackage{float}

\usepackage[colorlinks=true,citecolor=blue]{hyperref}

\usepackage{amsmath}
\usepackage{amsfonts}
\usepackage{bm}
\usepackage{xparse}
\usepackage{xpatch}
\usepackage{environ}
\usepackage{subcaption}
\usepackage{titletoc}

\makeatletter
\newcommand{\removeParBefore}{\ifvmode\vspace*{-\baselineskip}\setlength{\parskip}{0ex}\fi}
\newcommand{\removeParAfter}{\@ifnextchar\par\@gobble\relax}
\newcommand{\eq}{\begingroup\removeParBefore\endlinechar=32 \eqinner}
\newcommand{\eqinner}[2][aligned]{\endlinechar=32%
\begin{gather}\begin{#1}#2\end{#1}\end{gather}\endgroup\removeParAfter}
\makeatother

\DeclareDocumentCommand{\p}{ D<>{p} D<>{} r() }{
\def\content{#3}\patchcmd{\content}{|}{\;#2\vert\;}{}{}
\ensuremath{#1 #2(\content #2)}}

\DeclareDocumentCommand{\P}{ D<>{P} D<>{\big} r() }{
\def\content{#3}\patchcmd{\content}{|}{\;#2\vert\;}{}{}
\ensuremath{\operatorname{#1}#2(\content #2)}}

\DeclareDocumentCommand{\E}{ D<>{E} E{_}{{}} D<>{\big} r[] }{
\def\content{#4}
\ensuremath{\operatorname{#1}_{#2}#3[\content #3]}}

\DeclareDocumentCommand{\D}{ D<>{D} D<>{\big} r[] }{
\def\content{#3}\patchcmd{\content}{||}{\;#2\|\;}{}{}
\ensuremath{\operatorname{#1}\!#2[\content #2]}}

\NewDocumentCommand{\Uni}{ r() }{\P<U>](#1)}
\NewDocumentCommand{\Nor}{ r() }{\P<Normal>](#1)}
\NewDocumentCommand{\Cat}{ r() }{\P<Cat>](#1)}
\NewDocumentCommand{\Bin}{ r() }{\P<Bin>](#1)}
\NewDocumentCommand{\Bet}{ r() }{\P<Beta>](#1)}
\NewDocumentCommand{\Ber}{ r() }{\P<Bernoulli>(#1)}
\NewDocumentCommand{\Dir}{ r() }{\P<Dir>(#1)}

\DeclareDocumentCommand{\KL}{ D<>{\big} r[] }{\D<KL><#1>[#2]}
\DeclareDocumentCommand{\H}{ D<>{\big} r[] }{\E<H><#1>[#2]}
\DeclareDocumentCommand{\I}{ D<>{\big} r[] }{\E<I><#1>[#2]}

\newcommand{\norm}[1]{\left\lVert\,#1\,\right\rVert}
\newcommand\method{TECO}

\usepackage[accepted]{icml2023}

\usepackage{amsmath}
\usepackage{amssymb}
\usepackage{mathtools}
\usepackage{amsthm}

\usepackage[capitalize,noabbrev,nameinlink]{cleveref}

\definecolor{mydarkblue}{rgb}{0.0,0.1,0.5}
\hypersetup{%
colorlinks=true,
linkcolor=mydarkblue,
citecolor=mydarkblue,
filecolor=mydarkblue,
urlcolor=mydarkblue}

\theoremstyle{plain}

\theoremstyle{definition}

\theoremstyle{remark}

\usepackage[textsize=tiny]{todonotes}

\icmltitlerunning{Temporally Consistent Transformers for Video Prediction}

\begin{document}

\twocolumn[
\icmltitle{Temporally Consistent Transformers for Video Generation}



\icmlsetsymbol{equal}{*}

\begin{icmlauthorlist}
\icmlauthor{Wilson Yan}{berkeley}
\icmlauthor{Danijar Hafner}{toronto,deepmind}
\icmlauthor{Stephen James}{berkeley,dyson}
\icmlauthor{Pieter Abbeel}{berkeley}
\end{icmlauthorlist}

\icmlaffiliation{berkeley}{UC Berkeley}
\icmlaffiliation{toronto}{University of Toronto}
\icmlaffiliation{deepmind}{DeepMind}
\icmlaffiliation{dyson}{Dyson Robotics Lab}

\icmlcorrespondingauthor{Wilson Yan}{wilson1.yan@berkeley.edu}

\icmlkeywords{Machine Learning, ICML}

\vskip 0.3in
]



\printAffiliationsAndNotice{}  

\begin{abstract}
To generate accurate videos, algorithms have to understand the spatial and temporal dependencies in the world.
Current algorithms enable accurate predictions over short horizons but tend to suffer from temporal inconsistencies. When generated content goes out of view and is later revisited, the model invents different content instead.
Despite this severe limitation, no established benchmarks on complex data exist for rigorously evaluating video generation with long temporal dependencies.
In this paper, we curate 3 challenging video datasets with long-range dependencies by rendering walks through 3D scenes of procedural mazes, Minecraft worlds, and indoor scans.
We perform a comprehensive evaluation of current models and observe their limitations in temporal consistency.
Moreover, we introduce the Temporally Consistent Transformer (TECO), a generative model that substantially improves long-term consistency while also reducing sampling time.
By compressing its input sequence into fewer embeddings, applying a temporal transformer, and expanding back using a spatial MaskGit, TECO outperforms existing models across many metrics.
Videos are available on the website: \url{https://wilson1yan.github.io/teco}
\end{abstract}




\begin{figure}[t]
\centering
\includegraphics[width=.8\linewidth]{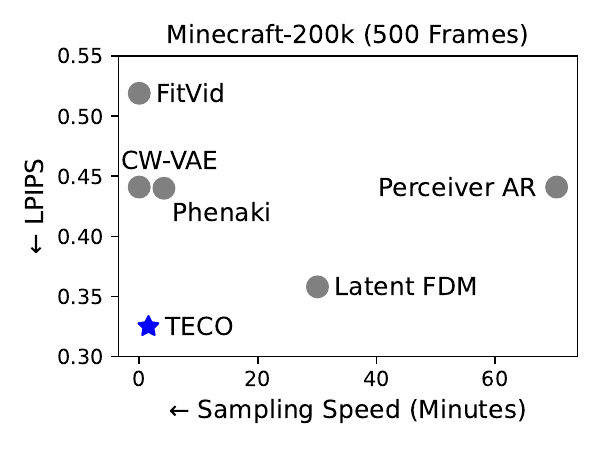}
\vspace*{-2ex}%
\caption{TECO generates temporally consistent videos of high fidelity (low LPIPS) over hundreds of frames while offering orders of magnitude faster sampling speed compared to previous video generation models.}
\label{fig:sampling_speed}
\vspace*{-3ex}%
\end{figure}
\begin{figure*}[t]
\centering
\includegraphics[width=\linewidth]{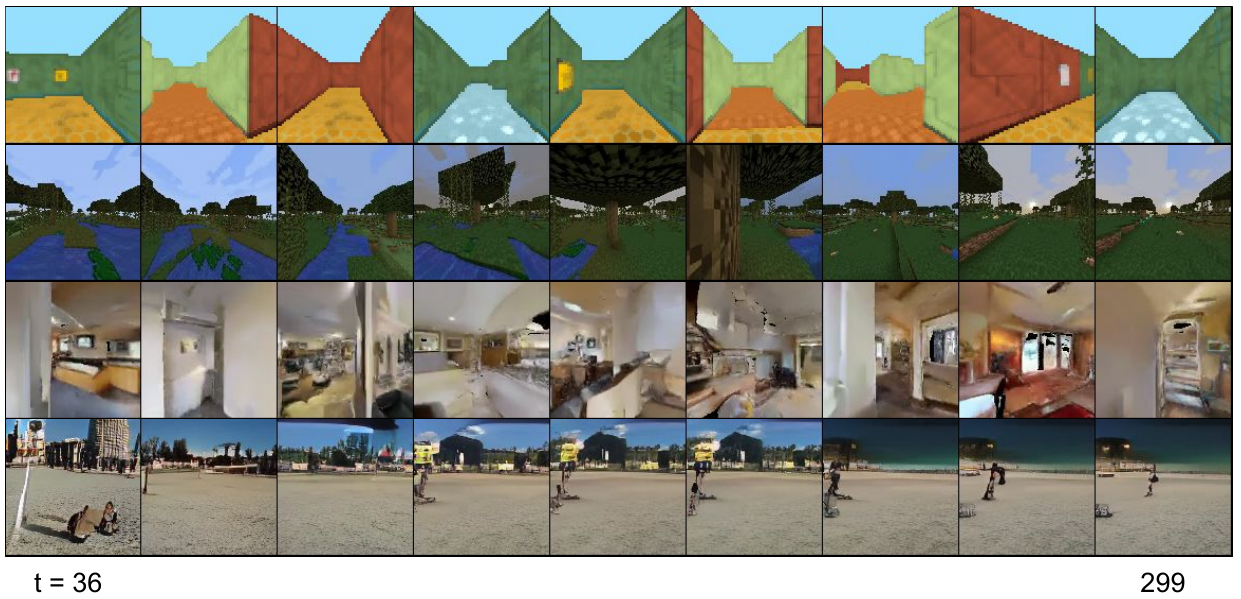}
\caption{TECO generates sharp and consistent video predictions for hundreds of frames on challenging datasets. The figure shows evenly spaced frames of the 264 frame predictions, after being conditioned on 36 context frames. From top to bottom, the datasets are are DMLab, Minecraft, Habitat, and Kinetics-600.}
\label{fig:teaser}
\end{figure*}

\section{Introduction}
Recent work in video generation has seen tremendous progress~\citep{ho2022video,clark2019adversarial,yan2021videogpt,le2021ccvs,ge2022long,tian2021good,luc2020transformation} in producing high-fidelity and diverse samples on complex video data, which can largely be attributed to a combination of increased computational resources and more compute efficient high-capacity neural architectures. However, much of this progress has focused on generating short videos, where models perform well by basing their predictions on only a handful of previous frames.

Video prediction models with short context windows can generate long videos in a sliding window fashion. While the resulting videos can look impressive at first sight, they lack temporal consistency. We would like models to predict temporally consistent videos --- where the same content is generated if a camera pans back to a previously observed location. On the other hand, the model should imagine a new part of the scene for locations that have not yet been observed, and future predictions should remain consistent to this newly imagined part of the scene. 

Prior work has investigated techniques for modeling long-term dependencies, such as temporal hierarchies~\citep{saxena2021clockwork} and strided sampling with frame-wise interpolation~\citep{ge2022long,hong2022cogvideo}. Other methods train on sparse sets of frames selected out of long videos~\citep{harvey2022flexible,skorokhodov2021stylegan,clark2019adversarial,saito2018tganv2,yu2022generating}, or model videos via compressed representations~\citep{yan2021videogpt,rakhimov2020latent,le2021ccvs,seo2022harp,gupta2022maskvit,walker2021predicting}. Refer to \Cref{section:related_work} for more detailed discussion on related work.

Despite this progress, many methods still have difficulty scaling to datasets with many long-range dependencies. While Clockwork-VAE~\citep{saxena2021clockwork} trains on long sequences, it is limited by training time (due to recurrence) and difficult to scale to complex data. On the other hand, transformer-based methods over latent spaces~\citep{yan2021videogpt} scale poorly to long videos due to quadratic complexity in attention, with long videos containing tens of thousands of tokens. Methods that train on subsets of tokens are limited by truncated backpropagation through time~\citep{hutchins2022block,rae2019compressive,dai2019transformer} or naive temporal operations~\citep{hawthorne2022general}.

In addition, there generally do not exist benchmarks for properly evaluating temporal consistency in video generation methods, where prior works either focus on generating long videos where short-term dependencies are sufficient for accurate prediction~\cite{ge2022long,skorokhodov2021stylegan} and/or rely on metrics such as FVD~\citep{unterthiner2019fvd} which are more sensitive to image fidelity rather than capture long-range temporal dependencies.


\begin{figure*}
    \includegraphics[width=\linewidth]{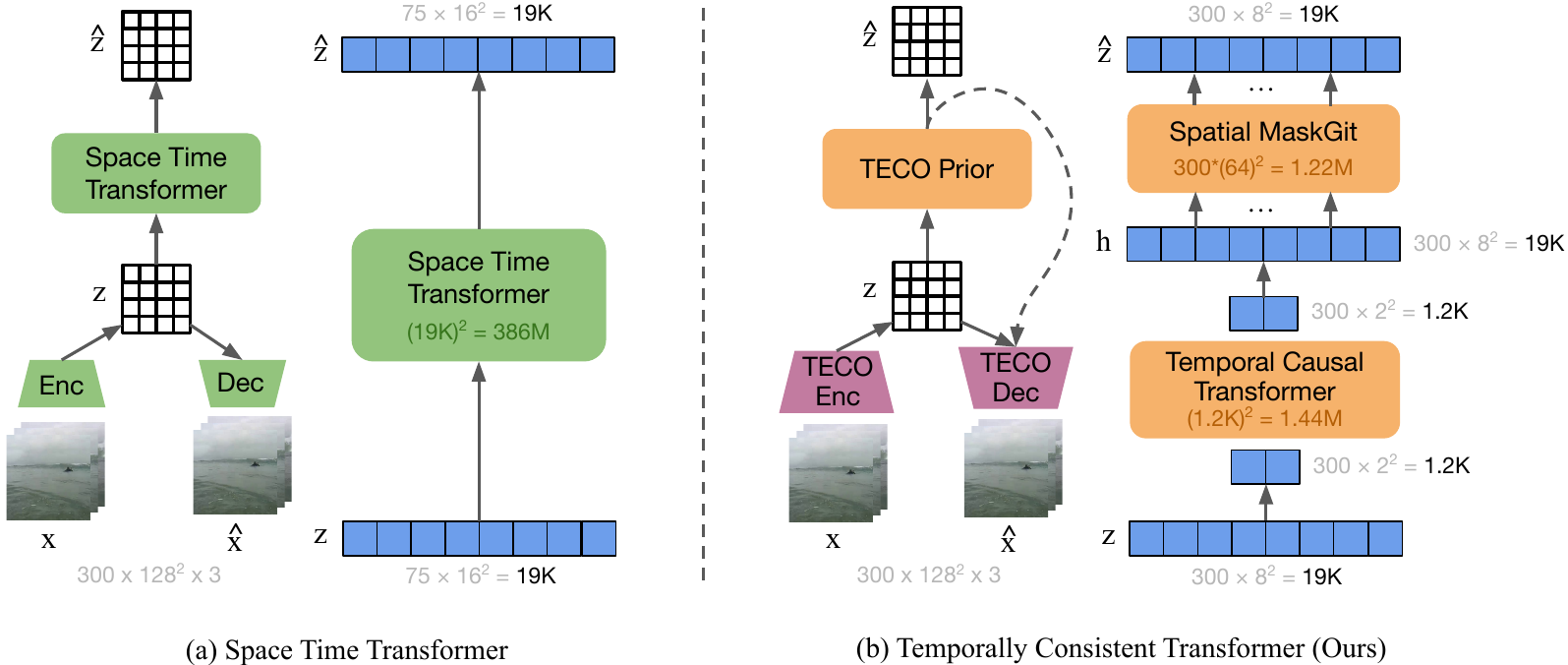}
    \caption{The architectural design of TECO. (a) Prior work on video generation models over VQ codes adopt a single spatio-temporal transformer over all codes. This is prohibitive when scaling to long sequences due to the quadratic complexity of attention. (b) We propose a novel and efficient architecture that aggressively downsamples in space before feeding into a temporal transformer, and then expands back out with a spatial MaskGit that is separately applied per frame. In the figure, the transformer blocks show the number of attention links. On training sequences of $300$ frames, TECO sees orders of magnitude more efficiency over existing models, allowing the use of larger models for a given compute budget.}
    \label{fig:arch2}
\end{figure*}

In this paper, we introduce a set of novel long-horizon video generation benchmarks, as well as corresponding evaluation metrics to better capture temporal consistency. In addition, we propose \textbf{Te}mporally \textbf{Co}nsistent Video Transformer (TECO), a vector-quantized latent dynamics model that effectively models long-term dependencies in a compact representation space using efficient transformers. The key contributions are summarized as follows: 
\begin{itemize}

    \item
    To better evaluate temporal consistency in video predictions, we propose 3 video datasets with long-range dependencies including metrics, generated from 3D scenes in DMLab~\citep{beattie2016deepmind}, Minecraft~\citep{guss2019minerl}, and Habitat~\citep{szot2021habitat,habitat19iccv}

    \item We benchmark SOTA video generation models on the datasets and analyze capabilities of each in learning long-horizon dependencies.

    \item We introduce TECO, an efficient and scalable video generation model that learns compressed representations to allow for efficient training and generation. We show that TECO has strong performance on a variety of difficult video prediction tasks, and is able to leverage long-term temporal context to generate high quality videos with consistency while maintaining fast sampling speed.
    
\end{itemize}

\section{Preliminaries}

\subsection{VQ-GAN}
VQ-GAN~\citep{esser2021taming, van2017neural} is an autoencoder that learns to compress data into discrete latents, consisting of an encoder $E$, decoder $G$, codebook $C$, and discriminator $D$. Given an image $x\in\mathbb{R}^{H\times W\times 3}$, the encoder $E$ maps $x$ to its latent representation $h\in\mathbb{R}^{H'\times W'\times D}$, which is quantized by nearest neighbors lookup in a codebook of embeddings $C = \{e_i\}^K_{i=1}$ to produce $z \in \mathbb{R}^{H'\times W' \times D}$. $z$ is fed through decoder $G$ to reconstruct $x$. A straight-through estimator~\citep{bengio2013estimating} is used to maintain gradient flow through the quantization step. The codebook optimizes the following loss:

\eq{
    \label{eq:vq}
    \mathcal{L}_{\mathrm{VQ}} &= \norm{\operatorname{sg}(h) - e}_2^2 + \beta\norm{h - \operatorname{sg}(e)}_2^2
}

where $\beta=0.25$ is a hyperparameter, and $e$ is the nearest-neighbors embedding from $C$. For reconstruction, VQ-GAN replaces the original $\ell_2$ loss with a perceptual loss~\citep{zhang2018unreasonable}, $\mathcal{L}_{\mathrm{LPIPS}}$. Finally, in order to encourage higher-fidelity samples, patch-level discriminator $D$ is trained to classify between real and reconstructed images, with:

\eq{
    \mathcal{L}_{\mathrm{GAN}} &= \log{D(x)} + \log(1 - D(\hat{x}))
}

Overall, VQ-GAN optimizes the following loss:
\eq{
    \min_{E,G,C}\max_{D} \,\,\, \mathcal{L}_{\mathrm{LPIPS}} + \mathcal{L}_{\mathrm{VQ}} + \lambda \mathcal{L}_{\mathrm{GAN}}
}
where $\lambda = \frac{\norm{\nabla_{G_L}\mathcal{L}_{\mathrm{LPIPS}}}_2}{\norm{\nabla_{G_L}\mathcal{L}_{\mathrm{GAN}}}_2 + \delta}$ is an adaptive weight, $G_L$ is the last decoder layer, $\delta = 10^{-6}$, and $\mathcal{L}_{LPIPS}$ is the same distance metric described in \citet{zhang2018unreasonable}.

\subsection{MaskGit}
MaskGit~\citep{chang2022maskgit} models distributions over discrete tokens, such as produced by a VQ-GAN. It generates images with competitive sample quality to autoregressive models at a fraction of the sampling cost by using a masked token prediction objective during training. Formally, we denote $z \in \mathbb{Z}^{H\times W}$ as the discrete latent tokens representing an image. For each training step, we uniformly sample $t\in [0, 1)$ and randomly generate a mask $m\in \{0,1\}^{H\times W}$ with $N=\lceil \gamma HW\rceil$ masked values, where $\gamma = \cos\left(\frac{\pi}{2}t\right)$. Then, MaskGit learns to predict the masked tokens with the following objective
\eq{
    \mathcal{L}_{\mathrm{mask}} = -\E_{z\in\mathcal{D}}[\log \p(z|z\odot m)].
}
During inference, because MaskGit has been trained to model any set of unconditional and conditional probabilities, we can sample any subset of tokens per sampling iteration. \cite{chang2022maskgit} introduces a confidence-based sampling mechanism whereas other work~\citep{lee2022draft} proposes an iterative sample-and-revise approach.

\section{\method{}}
We present \textbf{Te}mporally \textbf{Co}nsistent Video Transformer (\method{}), a video generation model that more efficiently scales to training on longer horizon videos. 



\subsection{Architectural Overview} 
Our proposed framework is shown in \cref{fig:arch2}, where $x_{1:T}$ consists of a sequence of video frames. Our primary innovation centers around designing a more efficient architecture that can scale to long sequences. Prior SOTA methods~\citep{yan2021videogpt,ge2022long,villegas2022phenaki} over VQ-codes all train a single spatio-temporal transformer to model every code, however, this becomes prohibitively expensive with sequences containing tens of thousands of tokens. On the other hand, these models have shown to be able to learn highly multi-modal distributions and scale well on complex video. As such, we design the TECO architecture with the intention to retain its high-capacity scaling properties, while ensuring orders of magnitude more efficient training and inference. In the following sections, we motivate each component for our model, with several specific design choices to ensure efficiency and scalability. TECO consists of four components:

\eq{
    &\text{Encoder:}\quad &&z_t = E(x_t, x_{t-1})\qquad \\
    &\text{Temporal Transformer:} &&h_t = H(z_{\leq t}) \\
    &\text{Spatial MaskGit:}\quad && \p(z_t\mid h_{t-1}) \\
    &\text{Decoder:} && \p(x_t\mid z_t,h_{t-1})
}

\textbf{Encoder}\quad
We can achieve compressed representations by leveraging spatio-temporal redundancy in video data. To do this, we learn a CNN encoder $z_t = E(x_t, x_{t-1})$ which encodes the current frame $x_t$ conditioned on the previous frame by channel-wise concatenating $x_{t-1}$, and then quantizes the output using codebook $C$ to produce $z_t$. We apply the VQ loss defined in \cref{eq:vq} per timestep. In addition, we $\ell_2$-normalize the codebook and embeddings to encourage higher codebook usage~\citep{yu2021vector}. The first frame is concatenated with zeros and does not quantize $z_1$ to prevent information loss. 


\textbf{Temporal Transformer}\quad
Compressed, discrete latents are more lossy and tend to require higher spatial resolutions compared to continuous latents. Therefore, before modeling temporal information, we apply a single strided convolution to downsample each discrete latent $z_t$, where visually simpler datasets allow for more downsampling and visually complex datasets require less downsampling. Afterwards, we learn a large transformer to model temporal dependencies, and then apply a transposed convolution to upsample our representation back to the original resolution of $z_t$. In summary, we use the following architecture:

\eq{
    h_t &= H(z_{<t}) = \text{ConvT}(\text{Transformer}(\text{Conv}(z_{<t})))
}

\textbf{Decoder}\quad
The decoder is an upsampling CNN that reconstructs $\hat{x_t} = D(z_t, h_t)$, where $z_t$ can be interpreted as the posterior of timestep $t$, and $h_t$ the output of the temporal transformer which summarizes information from previous timesteps. $z_t$ and $h_t$ are concatenated channel-wise and into the decoder. Together with the encoder, the decoder optimizes the following cross entropy reconstruction loss

\eq{
    \mathcal{L}_{\mathrm{recon}} &= -\textstyle\frac{1}{T}\sum_{t=1}^T \log p(x_t \mid z_t, h_t).
}

which encourages $z_t$ features to encode relative information between frames since the temporal transformer output $h_t$ aggregates information over time, learning more compressed codes for efficient modeling over longer sequences.

\textbf{Spatial MaskGit}\quad
Lastly, we use a MaskGit~\citep{chang2022maskgit} to model the prior, $p(z_t \mid h_t)$. We show that using a MaskGit prior allows for not just faster but also higher quality sampling compared to an autoregressive prior. During every training iteration, we follow prior work to sample a random mask $m_t$ and optimize

\eq{
    \mathcal{L}_{\mathrm{prior}} = -\textstyle\frac{1}{T}\sum_{t=1}^T \log\p(z_t \mid z_t \odot m_t).
}

where $h_t$ is concatenated channel-wise with masked $z_t$ to predict the masked tokens. During generation, we follow \citet{lee2022draft}, where we initially generate each frame in chunks of 8 at a time and then go through 2 revise rounds of re-generating half the tokens each time.

\textbf{Training Objective}\quad
The final objective is the following:
\begin{align}
    \mathcal{L}_{\mathrm{TECO}} &= \mathcal{L}_{\mathrm{VQ}} + \mathcal{L}_{\mathrm{recon}} + \mathcal{L}_{\mathrm{prior}}
\end{align}


\subsection{DropLoss} 

\begin{figure}[h]
    \centering
    \includegraphics[width=0.65\linewidth]{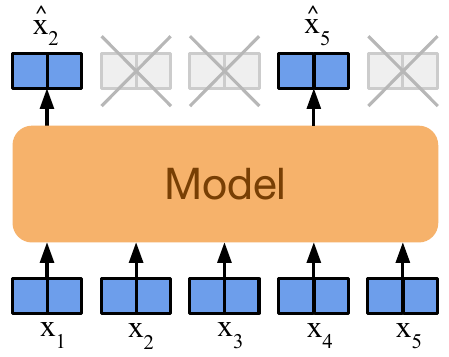}
    \vspace*{-2ex}
    \caption{DropLoss improves training scalability on longer sequences by only computing the loss on a random subset of time indices for each training iteration. For TECO, we do not need to compute the decoder and MaskGit for dropped out timesteps.}
    \label{fig:drop_loss}
    \vspace*{-2ex}
\end{figure}

We propose DropLoss, a simple trick to allow for more scalable and efficient training (\cref{fig:drop_loss}). Due to its architecture design, TECO can be separated into two components: (1) learning temporal representations, consisting of the encoder and the temporal transformer, and (2) predicting future frames, consisting of the dynamics prior and decoder. We can increase training efficiency by dropping out random timesteps that are not decoded and thus omitted from the reconstruction loss. For example, given a video of T frames, we compute $h_t$ for all $t \in \{1,\dots, T\}$, and then compute the losses $\mathcal{L}_{\mathrm{prior}}$ and $\mathcal{L}_{\mathrm{recon}}$ for only 10\% of the indices. Because random indices are selected each iteration, the model still needs to learn to accurately predict all timesteps. This reduces training costs significantly because the decoder and dynamics prior require non-trivial computations. DropLoss is applicable to both a wide class of architectures and to tasks beyond video prediction.

\begin{figure}[h]
    \centering
    \includegraphics[width=0.9\linewidth]{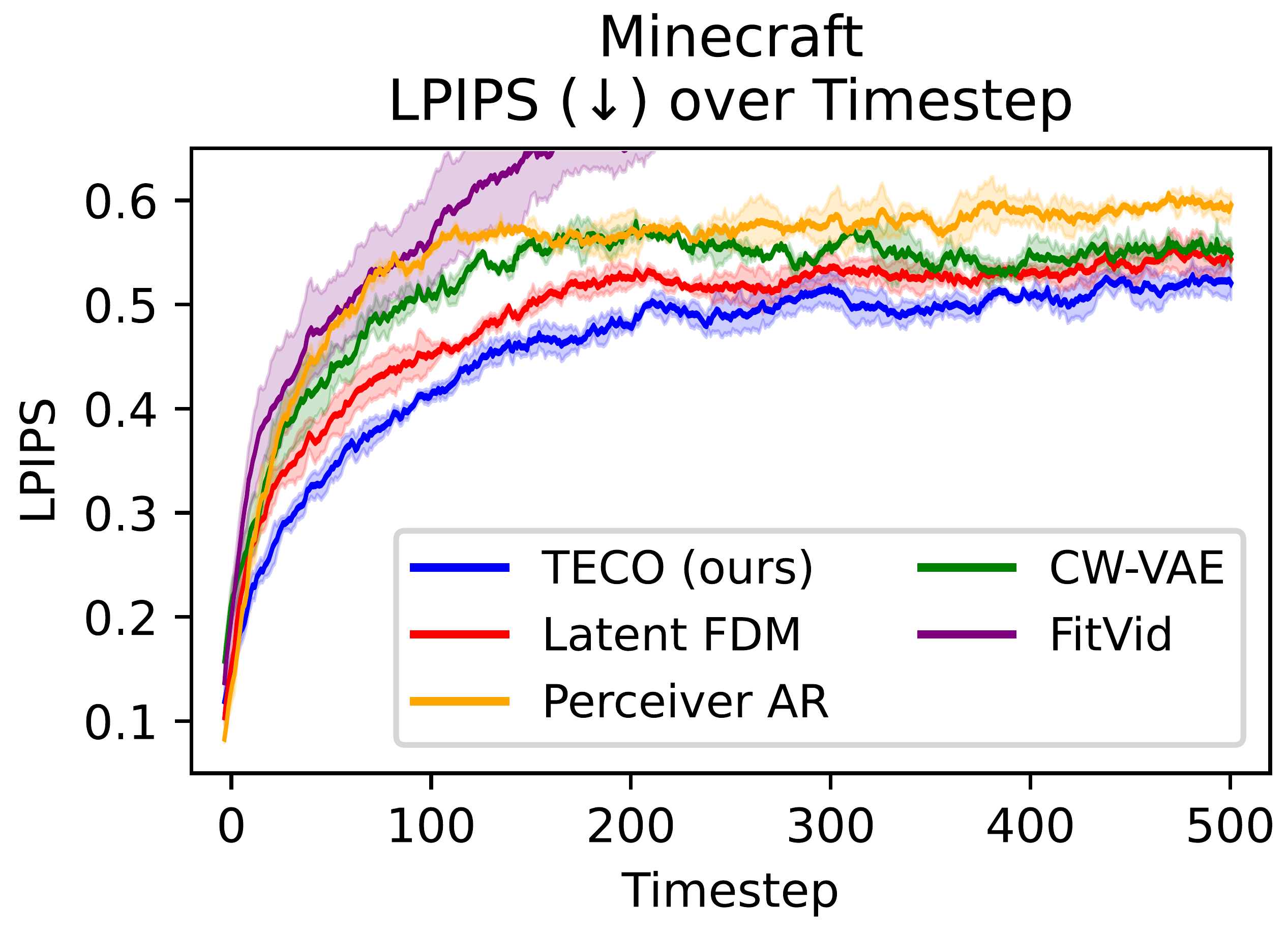}
    \vspace*{-2ex}
    \caption{Quantitative comparisons between TECO and baseline methods in long-horizon temporal consistency, showing LPIPS between generated and ground-truth frames for each timestep. Timestep 0 corresponds to the first predicted frame (conditioning frames are not included in the plot). Our method is able to remain more temporally consistent over hundreds of timesteps of prediction compared to SOTA models.}
    \label{fig:main_exp}
    \vspace*{-2ex}
\end{figure}

\section{Experiments}
\subsection{Datasets}
We introduce three challenging video datasets to better measure long-range consistency in video prediction, centered around 3D environments in DMLab~\citep{beattie2016deepmind}, Minecraft~\citep{guss2019minerl}, and Habitat~\citep{habitat19iccv}, with videos of agents randomly traversing scenes of varying difficulty. These datasets require video prediction models to re-produce observed parts of scenes, and newly generate unobserved parts. In contrast, many existing video benchmarks do not have strong long-range dependencies, where a model with limited context is sufficient. Refer to \Cref{section:dataset_details} for further details on each dataset.

\textbf{DMLab-40k}\quad DeepMind Lab is a simulator that procedurally generates random 3D mazes with random floor and wall textures. We generate 40k action-conditioned $64\times 64$ videos of $300$ frames of an agent randomly traversing $7\times 7$ mazes by choosing random points in the maze and navigating to them via the shortest path. We train all models for both action-conditioned and unconditional prediction (by periodically masking out actions) to enable both types of generations. We further discuss the use cases of both action and unconditional models in \Cref{section:evaluation}.

\textbf{Minecraft-200k}\quad This popular game features procedurally generated 3D worlds that contain complex terrain such as hills, forests, rivers, and lakes. We collect 200k action-conditioned videos of length $300$ and resolution $128 \times 128$ in Minecraft's marsh biome. The player iterates between walking forward for a random number of steps and randomly rotating left or right, resulting in parts of the scene going out of view and coming back into view later. We train action-conditioned for all models for ease of interpreting and evaluating, though it is generally easy for video models to unconditionally learn these discrete actions.

\textbf{Habitat-200k}\quad Habitat is a simulator for rendering trajectories through scans of real 3D scenes. We compile $\sim$1400 indoor scans from HM3D~\citep{ramakrishnan2021habitat}, MatterPort3D~\citep{chang2017matterport3d}, and Gibson~\citep{xia2018gibson} to generate 200k action-conditioned videos of $300$ frames at a resolution of $128 \times 128$ pixels. We use Habitat's in-built path traversal algorithm to construct action trajectories that move our agent between randomly sampled locations. Similar to DMLab, we train all video models to perform both unconditional and action-conditioned prediction.

\textbf{Kinetics-600}\quad Kinetics-600~\citep{carreira2017quo} is a highly complex real-world video dataset, originally proposed for action recognition. The dataset contains $\sim$400k videos of varying length of up to 300 frames. We evaluate our method in the video prediction without actions (as they do not exist), generating 80 future frames conditioned on 20. In addition, we filter out videos shorter than 100 frames, leaving 392k videos that are split for training and evaluation. We use a resolution of $128 \times 128$ pixels. Although Kinetics-600 does not have many long-range dependencies, we evaluate our method on this dataset to show that it can scale to complex, natural video.

\begin{table*}[!t]
\centering
\vspace*{0ex}
\caption{Quantitative evaluation on all four datasets. Detailed results in \Cref{section:full_results}.}
\label{table:main}
\vspace*{0ex}
\begin{tabular}{@{}lccccccccc@{}}
\toprule
                                 & \multicolumn{4}{c}{DMLab}                                            & \multicolumn{4}{c}{Minecraft}                                              \\
Method                       & FVD $\downarrow$ & PSNR $\uparrow$ & SSIM $\uparrow$  & LPIPS $\downarrow$ & FVD $\downarrow$ & PSNR $\uparrow$ & SSIM $\uparrow$  & LPIPS $\downarrow$ \\ \midrule
FitVid                       & $176$            & $12.0$          & $0.356$          & $0.491$            & $956$            & $13.0$          & $0.343$          & $0.519$            \\
CW-VAE                       & $125$            & $12.6$          & $0.372$          & $0.465$            & $397$            & $13.4$          & $0.338$          & $0.441$            \\
Perceiver \rlap{AR}                 & $\hphantom{0}96$           & $11.2$          & $0.304$          & $0.487$            & $\hphantom{0}\textbf{76}$            & $13.2$          & $0.323$          & $0.441$            \\ 
Latent FDM  & $181$            & $17.8$          & $0.588$          & $0.222$            & $167$            & $13.4$          & $0.349$          & $0.429$            \\ 
TECO (ours) & $\hphantom{0}\mathbf{48}$  & $\mathbf{21.9}$ & $\mathbf{0.703}$ & $\mathbf{0.157}$   & $116$            & $\mathbf{15.4}$          & $\mathbf{0.381}$          & $\mathbf{0.340}$            \\
\bottomrule
\toprule
             & \multicolumn{4}{c}{Habitat}                                                & \multicolumn{4}{c}{Kinetics-600}                                           \\
Method       & FVD $\downarrow$ & PSNR $\uparrow$ & SSIM $\uparrow$  & LPIPS $\downarrow$ & FVD $\downarrow$ & PSNR $\uparrow$ & SSIM $\uparrow$  & LPIPS $\downarrow$ \\ \midrule
Perceiver AR & $164$            & $\mathbf{12.8}$ & $\mathbf{0.405}$ & $0.676$            & $\llap{1}022$           & $13.4$          & $0.310$          & $0.404$            \\
Latent FDM   & $433$            & $12.5$          & $0.311$          & $\mathbf{0.582}$   & $960$            & $13.2$          & $0.334$          & $0.413$            \\
TECO (ours)  & $\hphantom{0}\mathbf{73}$    & $\mathbf{12.8}$ & $0.363$          & $0.604$            & $\mathbf{799}$   & $\mathbf{13.8}$ & $\mathbf{0.341}$ & $\mathbf{0.381}$   \\ \bottomrule
\end{tabular}
\end{table*}

\begin{figure*}[ht!]
    \centering
    \vspace*{0ex}
    \includegraphics[width=0.85\linewidth]{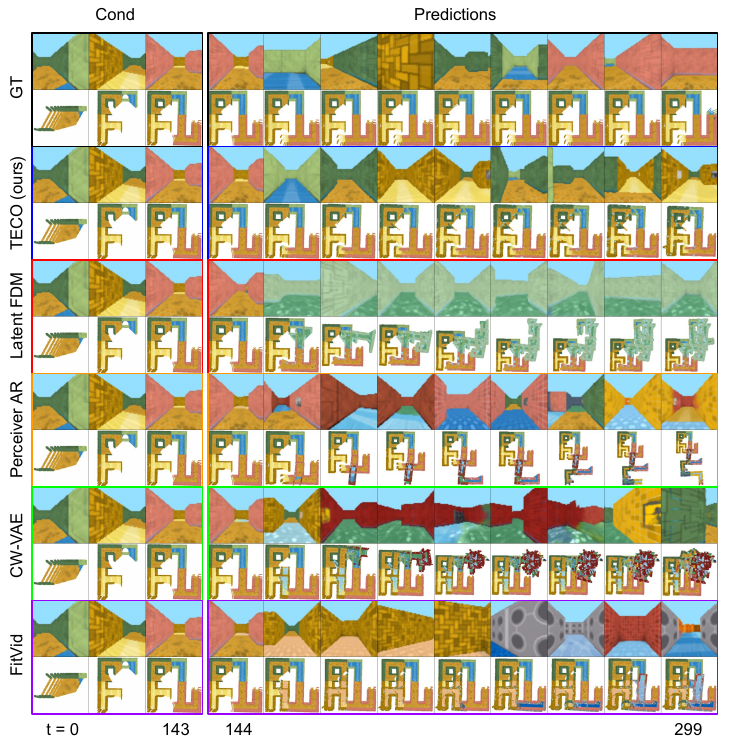}
    \vspace*{-2ex}
    \caption{3D visualization of predicted trajectories in DMLab for each model, generating 156 frames conditioned on 144. TECO is the only model that retain maze consistency with ground-truth, whereas baselines tend to extend out of the maze or create fictitious corridors that did not exist. \textbf{Video predictions use only the first-person RGB frames}. Refer to \Cref{section:dataset_dmlab_details} for more details on 3D evaluation. A video corresponding to this figure is available at: \url{https://wilson1yan.github.io/teco}.}
    \label{fig:dm_lab_3d_traj}
    \vspace*{0ex}
\end{figure*}

\subsection{Baselines}
We compare against SOTA baselines selected from several different families of models: latent-variable-based variational models, autoregressive likelihood models, and diffusion models. In addition, for efficiency, we train all models on VQ codes using a pretrained VQ-GAN for each dataset. For our diffusion baseline, we follow \cite{rombach2022high} and use a VAE instead of a VQ-GAN. Note that we do not have any GANs for our baselines, since to the best of our knowledge, there does not exist a GAN that trains on latent space instead of raw pixels, an important aspect for properly scaling to long video sequences.

\textbf{Space-time Transformers}\quad
We compare TECO to several variants of space-time transformers as depicted in \Cref{fig:arch2}: VideoGPT~\citep{yan2021videogpt} (autoregressive over space-time), Phenaki~\citep{villegas2022phenaki} (MaskGit over space-time full attention), MaskViT~\citep{gupta2022maskvit} (MaskGit over space-time axial attention), and Hourglass transformers~\citep{nawrot2021hierarchical} (hierarchical autoregressive over space-time). Note that we do not include the text-conditioning for Phenaki as it is irrelevant in our case. We only evaluate these models on DMLab, as \Cref{table:a_b_comp} and \Cref{table:main} show that Perceiver-AR (a space-time transformer with improvements specifically for learning long dependencies) is a stronger baseline.

\textbf{FitVid}\quad
FitVid~\citep{babaeizadeh2021fitvid} is a state-of-the-art variational video model based on CNNs and LSTMs that scales to complex video by leveraging efficient architectural design choices in its encoder and decoder.

\textbf{Clockwork VAE}\quad
CW-VAE \citep{saxena2021clockwork} is a variational video model that is designed to learn long-range dependencies through a hierarchy of latent variables with exponentially slower tick speeds for each new level. 

\textbf{Perceiver AR}\quad
We use Perceiver AR~\citep{hawthorne2022general} as our AR baseline over VQ-GAN discrete latents, which has been show to be an effective generative model that can efficiently incorporate long-range sequential dependencies. Conceptually, this baseline is similar to HARP~\citep{seo2022harp} with a Perceiver AR as the prior instead of a sparse transformer~\citep{child2019generating}. We choose Perceiver AR over other autoregressive baselines such as VideoGPT~\citep{yan2021videogpt} or TATS~\citep{ge2022long} primarily due to the prohibitive costs of transformers when applied to tens of thousands of tokens.

\textbf{Latent FDM}\quad
We train a Latent FDM model for our diffusion baseline. Although FDM~\citep{harvey2022flexible} is originally trained on pixel observations, we also train in latent space for a more fair comparison with our method and other baselines, as training on long sequences in pixel space is too expensive. We follow LDM~\citep{rombach2022high} to separately train an autoencoder to encode each frame into a set of continuous latents. 

\begin{table}[]
\centering
\caption{TECO substantially outperforms similar video generation models that use space-time transformers.}
\label{table:a_b_comp}
\begin{tabular}{@{}lllll@{}}
\toprule
Model                 & FVD$\downarrow$ & PSNR$\uparrow$ & SSIM$\uparrow$ & LPIPS$\downarrow$ \\ \midrule
TATS                  & 156             & 11.1           & 0.296          & 0.468             \\
Phenaki               & 725             & 11.0           & 0.202          & 0.474             \\
MaskViT               & 76              & 12.4           & 0.360          & 0.435             \\
Hourglass & 110             & 11.7           & 0.335          & 0.458             \\
TECO (ours)           & \textbf{48}     & \textbf{21.9}  & \textbf{0.703} & \textbf{0.157}    \\ \bottomrule
\end{tabular}
\vspace*{-2ex}
\end{table}

\subsection{Experimental Setup}
\textbf{Training}\quad
All models are trained for 1 million iterations under fixed compute budgets allocated for each dataset (measured in TPU-v3 days) on TPU-v3 instances ranging from v3-8 to v3-128 TPU pods (similar to 4 V100s to 64 V100s) with training times of roughly 3-5 days. For DMLab, Minecraft, and Habitat we train all models on full 300 frames videos, and 100 frames for Kinetics-600. Our VQ-GANs are trained on 8 A5000 GPUs, taking about 2-4 days for each dataset, and downsample all videos to $16\times 16$ grids of discrete latents per frame regardless of original video resolution. More details on exact hyperparameters and compute budgets for each dataset can be found in \Cref{section:hyperparams}.

\textbf{Evaluation}\quad
\label{section:evaluation}

Standard methods for evaluating video prediction quality (FVD~\citep{unterthiner2019fvd} or per-frame metrics PSNR~\citep{huynh2008scope}, SSIM~\citep{wang2004image}, and LPIPS~\citep{zhang2018unreasonable}) do not measure long-consistency well. FVD is more sensitive to image fidelity, and relies on an I3D network trained on short Kinetics-600 clips. Evaluations using PSNR, SSIM, and LPIPS generally require sampling hundreds of futures and compare the sample that most accurately matches ground-truth. However, this does not align well with the goal of temporal consistency, as we would like the model to deterministically re-generate observed parts of the environment, and not accidentally generate the correct future after many samples.

Therefore, we propose a modified evaluation metric using PSNR, SSIM, and LPIPS that better measures temporal consistency in video generation by leveraging sufficient conditioning. Intuitively, if a video model is conditioned with enough information, future predictions should be approximately deterministic, meaning that only one sample should be needed to expect an accurate match with ground-truth. In the case of 3D environments, we can approximately make generation deterministic by conditioning on past frames (after the model has already seen most of the 3D environment) and actions (to remove stochasticity of movement). As such, for DMLab, Minecraft, and Habitat, we condition on 144 past frames as well as actions, and measure PSNR, SSIM, and LPIPS with 156 future ground-truth frames. However, note that per-frame metrics only capture temporal consistency, and do not capture a video model's ability to model the stochasticity of video data. Therefore, we also compute FVD on 300 frame videos, conditioned on 36 frames (264 predicted frames). For Kinetics-600, we evaluate FVD on 100 frame videos, conditioned on 20 frames (80 predicted frames). We compute all metrics over batches of $256$ examples, averaged over 4 runs to make $1024$ total samples.

\vspace*{-.5ex}
\subsection{Benchmark Results}
\vspace*{-.1ex}
\textbf{DMLab \& Minecraft}\quad
\Cref{table:main} shows quantitative results on the DMLab and Minecraft datasets. TECO performs the best across all metrics for both datasets when training on the full 300 frame videos.  \Cref{fig:dm_lab_3d_traj} shows sample trajectories and 3D visualizations of the generated DMLab mazes, where TECO is able to generate more consistent 3D mazes. For both datasets, CW-VAE, FitVid, and Perceiver AR can produce sharp predictions, but do not model long-horizon context well, with per-frame metrics sharply dropping as the future prediction horizon increases as seen in \Cref{fig:horizon}. Latent FDM has consistent predictions, but high FVD most likely due to FVD being sensitive to high frequency errors.

\textbf{Habitat}\quad
\Cref{table:main}
shows results for our Habitat dataset. We only evaluate our strongest baselines, Perceiver AR and Latent FDM due to the need to implement model parallelism. Because of high complexity of Habitat videos, all models generally perform equally as bad in per-frame metrics. However, TECO has significantly better FVD. Qualitatively, Latent FDM quickly collapses to blurred predictions with poor sample quality, and Perceiver AR generates high quality frames, though less temporally consistent than TECO: agents in Habitat videos navigate to far points in the scene and back whereas Perceiver AR tends to generate samples where the agent constantly turns. TECO generates traversals of a scene that match the data distribution more closely.

\textbf{Kinetics-600}\quad
\Cref{table:main} shows FVD for predicting $80$ $128\times128$ frames conditioned on $20$ for Kinetics-600. Although Kinetics-600 does not have many long-range dependencies, we found that TECO is able to produce more stable generations that degrade slower by incorporating longer contexts. In contrast, Perceiver AR tends to degrade quickly, with Latent FDM performing in between. 

\textbf{Sampling Speed}\quad
\Cref{fig:main_exp} reports sampling speed for all models on Minecraft. We observed similar results for the different model sizes used on other datasets. FitVid and CW-VAE are both significantly faster that other methods, but have poor sample quality. On the other end, Perceiver AR and Latent FDM can produce high quality samples, but are 20-60x slower than TECO, which has comparably fast sampling speed while retaining high sample quality.

\vspace*{-.5ex}
\subsection{Ablations}
\vspace*{-1ex}
In this section, we perform ablations on various architectural decisions of our model. For simplicity, we evaluate our methods on short sequences of 16 frames from Something-Something-v2 (SSv2), as it provides insight into scaling our method on complex real-world data more similar to Kinetics-600 while being computationally cheaper to run. 

Details can be found in the Appendix, \Cref{table:abl_arch}. We demonstrate that using VQ-latent dynamics with a MaskGit prior outperforms other formulations for latent dynamics models such as variational methods. In addition, we show that conditional encodings learn better representations for video predictions. We also ablate the codebook size, showing that although there exists an optimal codebook size, it does not matter too much as along as there are not too many codes, which may the prior more difficult to learn. Lastly, we show the benefits of DropLoss, with up to 60\% faster training and a minimal increase in FVD. The benefits are greater for longer sequences, and allow video models to better account for long horizon context with little cost in performance.

\vspace*{-.5ex}
\subsection{Further Insights}
\vspace*{-1ex}
We highlight a few key experimental insights for designing long-horizon video generation models. Further details can be found in \Cref{section:tradeoff} and \Cref{section:metricstraining}.

\textbf{Trade-off between fidelity and learning long-range dependencies}\quad
Given a network with fixed capacity, there exists an inherent trade-off between generating high fidelity and temporally consistent videos. We find that long-horizon information can be prioritized through bottlenecking representations, whereas allocating more computation towards higher resolution representations encourages higher fidelity. Due to TECO learning more compact representations, it achieves a better trade-off between fidelity and temporal consistency compared to our baseline models, demonstrated by better PSNR / SSIM / LPIPS, in addition to FVD.

\textbf{Although frame quality saturates early-on, long-term consistency improves when training longer}\quad
During training, we observe an interesting phenomenon where short-horizon metrics tend to saturate earlier on during training, while long-horizon metrics continue to improve until end of training. We hypothesize that this may be due to the likelihood objective, where modeling bits from neighboring frames is easier than learning long-horizon bits scattered throughout the video. This finding motivates the use of an efficient video architecture for TECO, which can be trained for more gradient steps given a fixed computational budget.

\vspace*{-.5ex}
\section{Discussion}
\vspace*{-1ex}

We introduced TECO, an efficient video prediction model that leverages hundreds of frames of temporal context, as well as a comprehensive benchmark to evaluate long-horizon consistency. Our evaluation demonstrated that TECO accurately incorporates long-range context, outperforming SOTA baselines across a wide range of datasets. In addition, we introduce several difficult video datasets, which we hope make it easier to evaluate temporal consistency in future video prediction models. We identify several limitations as directions for future work:
\begin{itemize}
    \item Although we show that PSNR, SSIM, and LPIPS can be reliable metrics to measure consistency when video models are properly conditioned, there remains room for better evaluation metrics that provide a reliable signal as the prediction horizon grows, since new parts of a scene that are generated are unlikely to correlate with ground truth.
    
    \item Our focus was on learning a compressed tokens and an expressive prior, which we combined with a simple full attention transformer as the sequence model. Leveraging prior work on efficient sequence models~\citep{choromanski2020rethinking,wang2020linformer,zhai2021attention,gu2021efficiently,hawthorne2022general} would likely allow for further scaling. 
    
    \item We trained all models on top of pretrained VQ-GAN codes to reduce the data dimensionality. This compression step lets us train on longer sequences at a cost of reconstruction error, which causes noticeable artifacts in Kinetics-600, such as corrupted text and incoherent faces. Although TECO can train directly on pixels, a $\ell_2$ loss results in slightly blurry predictions. Training directly on pixels with diffusion or GAN losses would be promising.
  
\end{itemize}

\section{Acknowledgements}
This work was in part supported by Panasonic through BAIR Commons, Darpa RACER, the Hong Kong Centre for Logistics Robotics, and BMW. In addition, we thank teh TRC program (\url{https://sites.research.google/trc/about/}) and Cirrascale Cloud Services (\url{https://cirrascale.com/}) for providing compute resources.

\clearpage
\bibliography{example_paper}
\bibliographystyle{icml2023}

\newpage
\appendix
\onecolumn
\counterwithin{figure}{section}
\counterwithin{table}{section}

\section*{Table of Contents}
\hyperref[section:samplingprocess]{{\large A\quad Sampling Process}}\\[1ex]
\hyperref[section:samples]{{\large B\quad Samples}}\\[1ex]
\hyperref[section:perfhorizon]{{\large C\quad Performance versus Horizon}} \\[1ex]
\hyperref[section:perfseqlength]{{\large D\quad Performance versus Training Sequence Length}} \\[1ex]
\hyperref[section:samplingtime]{{\large E\quad Sampling Time}} \\[1ex]
\hyperref[section:ablations]{{\large F\quad Ablations}} \\[1ex]
\hyperref[section:metricstraining]{{\large G\quad Metrics During Training}} \\[1ex]
\hyperref[section:compression]{{\large H\quad High Quality Spatio-Temporal Compression}} \\[1ex]
\hyperref[section:tradeoff]{{\large I\quad Trade-off Between Fidelity and Learning Long-Range Dependencies}} \\[1ex]
\hyperref[section:full_results]{{\large J\quad Full Experimental Results}} \\[1ex]
\hyperref[section:scaling]{{\large K\quad Scaling Results}} \\[1ex]
\hyperref[section:related_work]{{\large L\quad Related Work}} \\[1ex]
\hyperref[section:dataset_details]{{\large M\quad Dataset Details}} \\[1ex]
\hyperref[section:hyperparams]{{\large N\quad Hyperparameters}} \\

\clearpage

\section{Sampling Process}
\label{section:samplingprocess}
Given a sequence of conditioning frames, $o_1, \dots, o_t$, we encode each frame using the pretrained VQ-GAN to produce $x_1, \dots, x_t$, and then use the conditional encoder to compute $z_1, \dots, z_t$. In order to generate the next frame, we use the temporal transformer to compute $h_t$, and feed it into the MaskGit dynamics prior to predict $\hat{z}_{t+1}$. Let $z_{t+1} = \hat{z}_{t+1}$ and feed it through the temporal tranformer and MaskGit to predict $\hat{z}_{t+2}$. We repeat this process until the entire trajectory is predicted, $\hat{z}_{t+1}, \dots, \hat{z}_T$. In order to decode back into frames, we first decode into the VQ-GAN latents, and then decode to RGB using the VQ-GAN decoder. Note that generation can be completely done in latent space, and rendering back to RGB can be done in parallel over time once the latents for all timesteps are computed.
\clearpage

\section{Samples}
\label{section:samples}

\subsection{DMLab}
\label{section:samplesdmlab}

\begin{figure}[H]
    \centering
    \includegraphics[width=\linewidth]{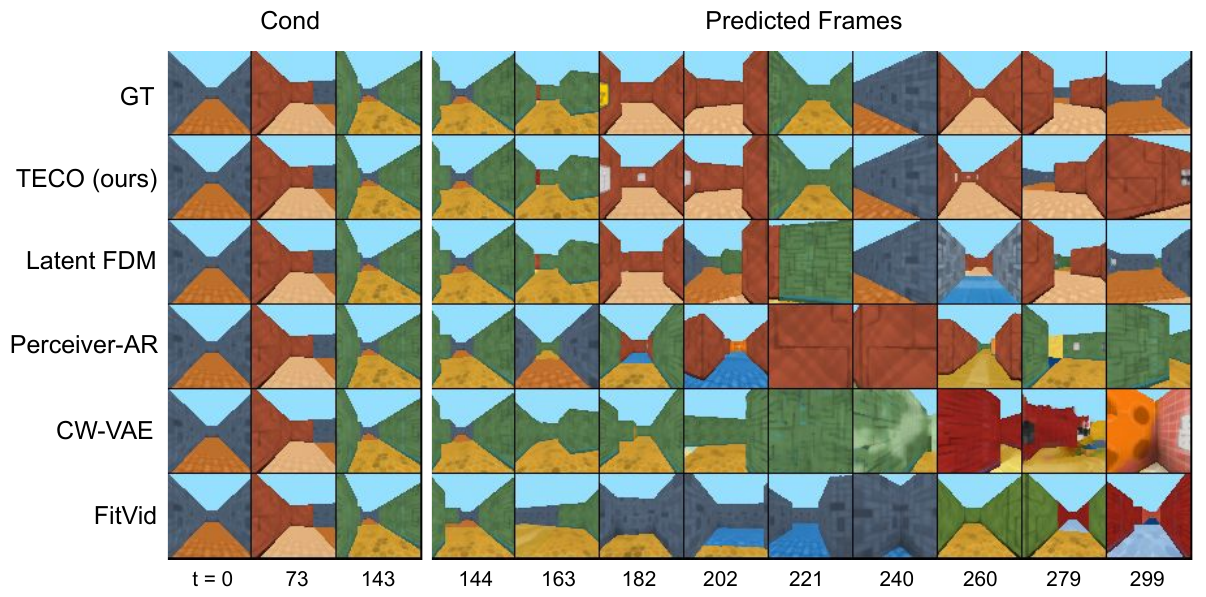}
    \caption{156 frames generated conditioned on 144 (action-conditioned)}
\end{figure}

\begin{figure}[H]
    \centering
    \includegraphics[width=\linewidth]{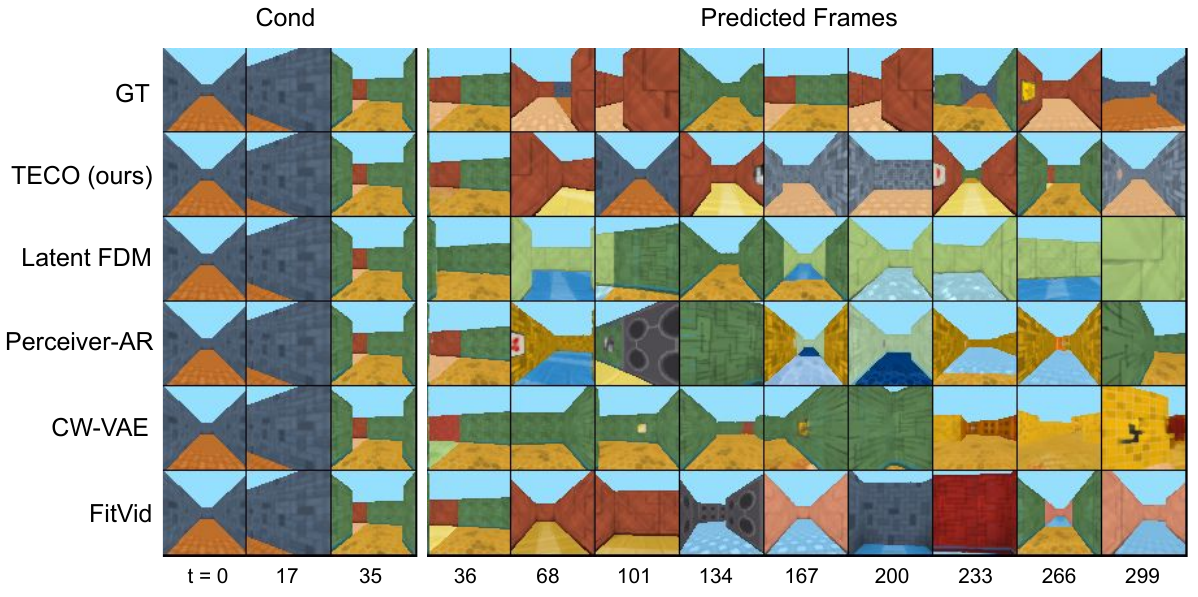}
    \caption{264 frames generated conditioned on 36 (\textbf{no} action-conditioning)}
\end{figure}

\clearpage

\begin{figure}[H]
    \centering
    \includegraphics[width=\linewidth]{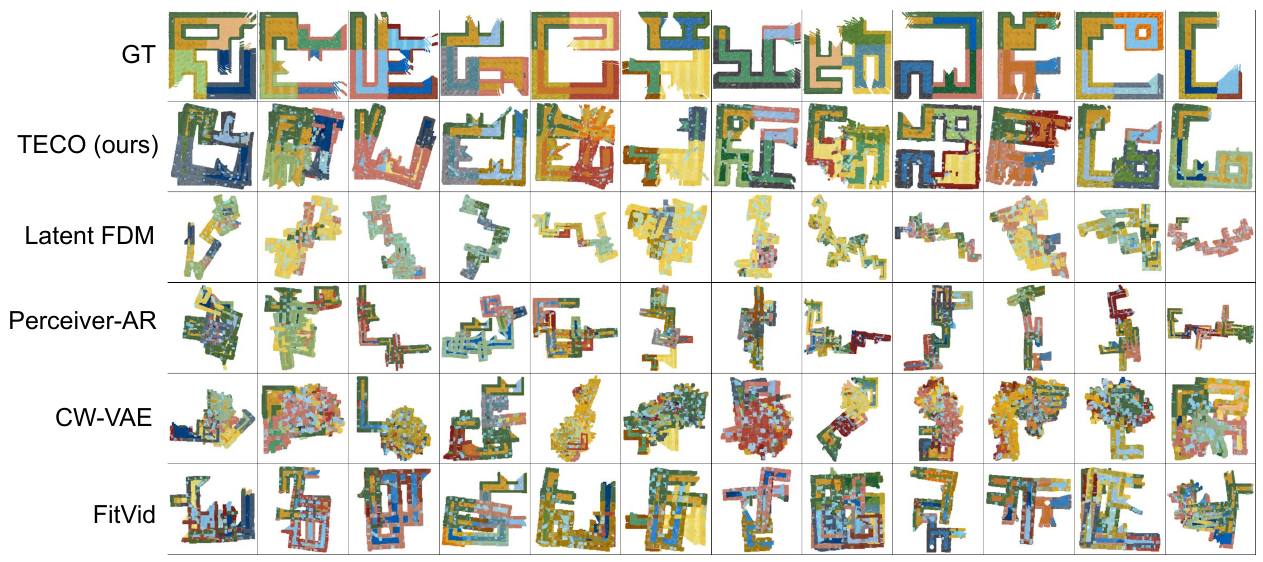}
    \caption{3D visualizations of the resulting generated DMLab mazes}
    \label{fig:dmlab_3d_more}
\end{figure}

\clearpage

\clearpage

\subsection{Minecraft}
\label{section:samplesminecraft}
\begin{figure}[H]
    \centering
     \includegraphics[width=\linewidth]{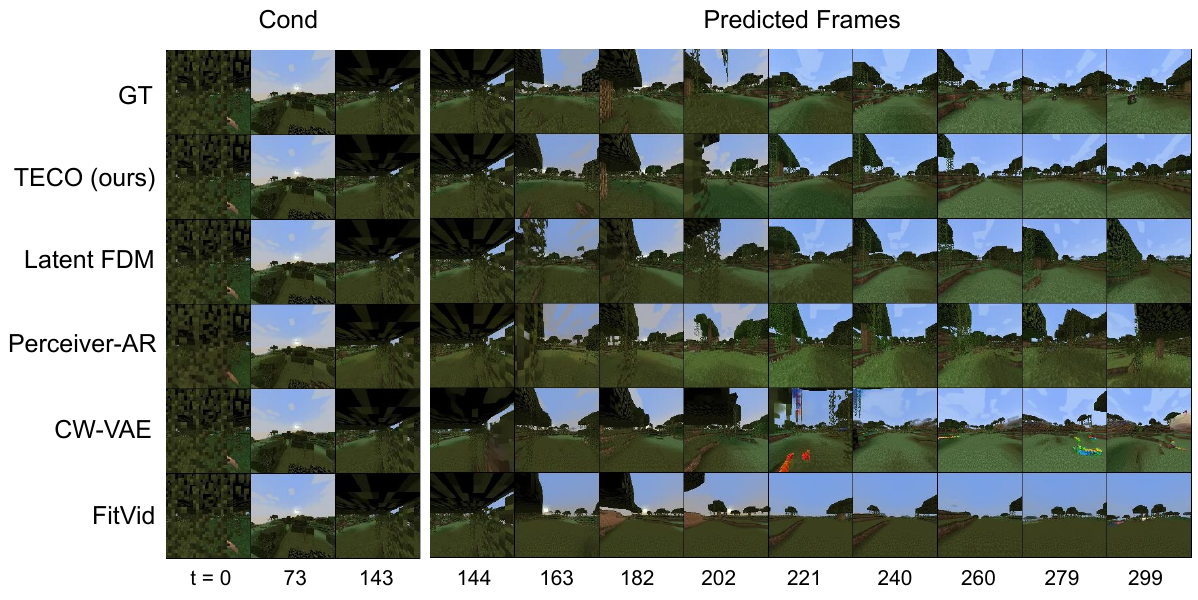}
    \caption{156 frames generated conditioned on 144 (action-conditioned)}
\end{figure}

\begin{figure}[H]
    \centering
     \includegraphics[width=\linewidth]{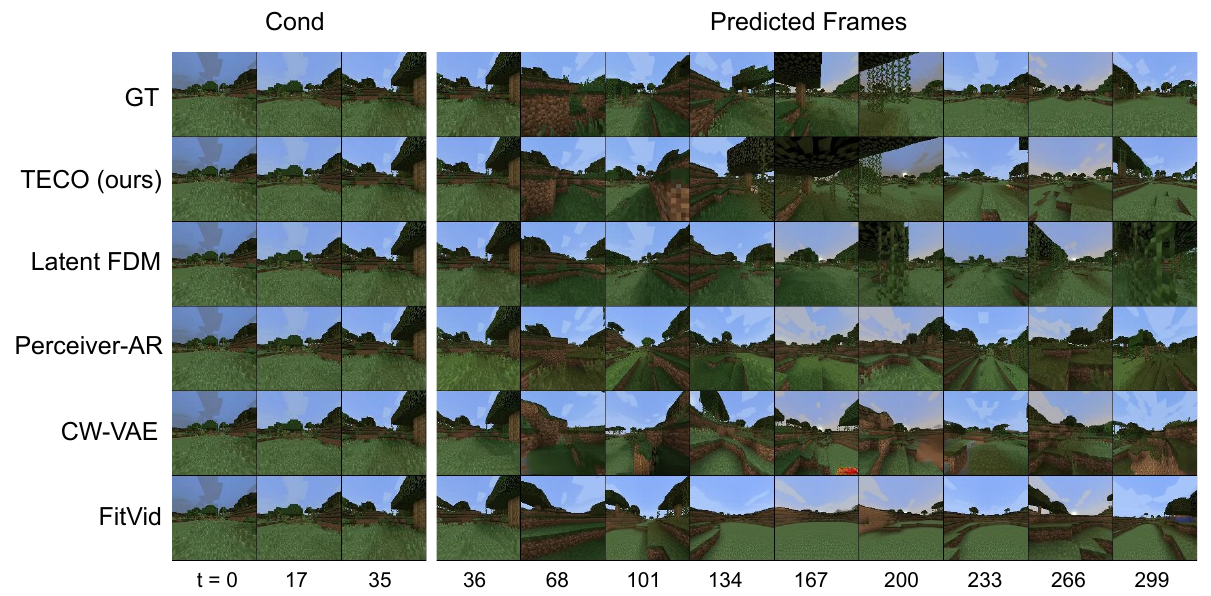}
    \caption{264 frames generated conditioned on 36 (action-conditioned)}
\end{figure}

\clearpage

\subsection{Habitat}
\label{section:sampleshabitat}

\begin{figure}[H]
    \centering
    \includegraphics[width=\linewidth]{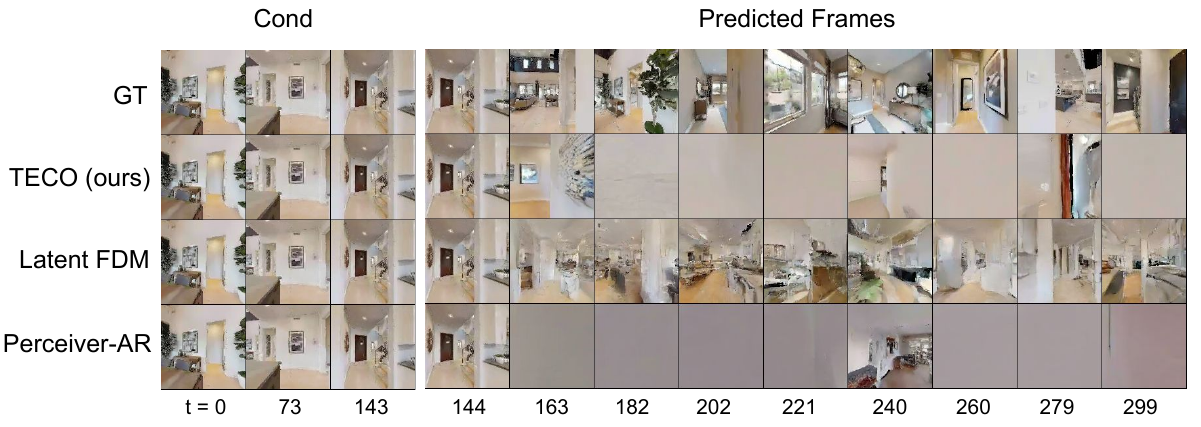}
    \caption{156 frames generated conditioned on 144 (action-conditioned)}
\end{figure}

\begin{figure}[H]
    \centering
    \includegraphics[width=\linewidth]{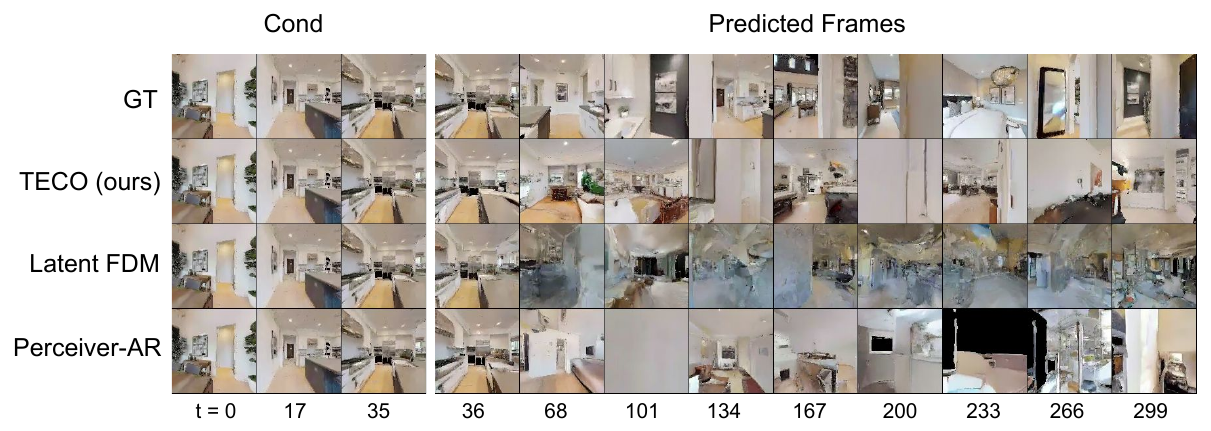}
    \caption{264 frames generated conditioned on 36 (\textbf{no} action-conditioning)}
\end{figure}

\clearpage

\subsection{Kinetics-600}
\label{section:sampleskinetics}

\begin{figure}[H]
    \centering
     \includegraphics[width=\linewidth]{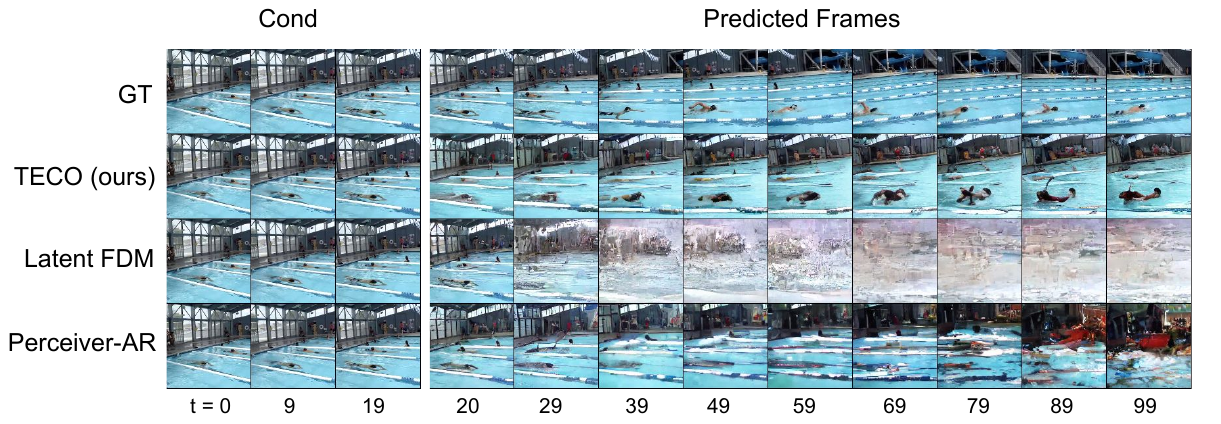}
    \caption{80 frames generated conditioned on 20 (no top-k sampling)}
\end{figure}

\begin{figure}[H]
    \centering
    \includegraphics[width=\linewidth]{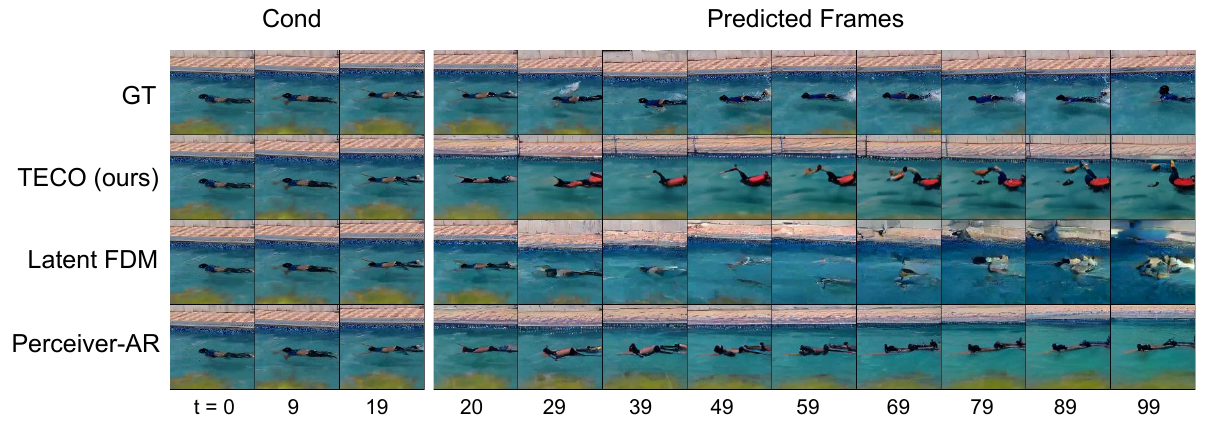}
    \caption{80 frames generated conditioned on 20 (with top-k sampling)}
\end{figure}

\clearpage

\section{Performance versus Horizon}
\label{section:perfhorizon}

\begin{figure}[H]
\begin{subfigure}[b]{\linewidth}%
\rlap{\includegraphics[width=\linewidth]{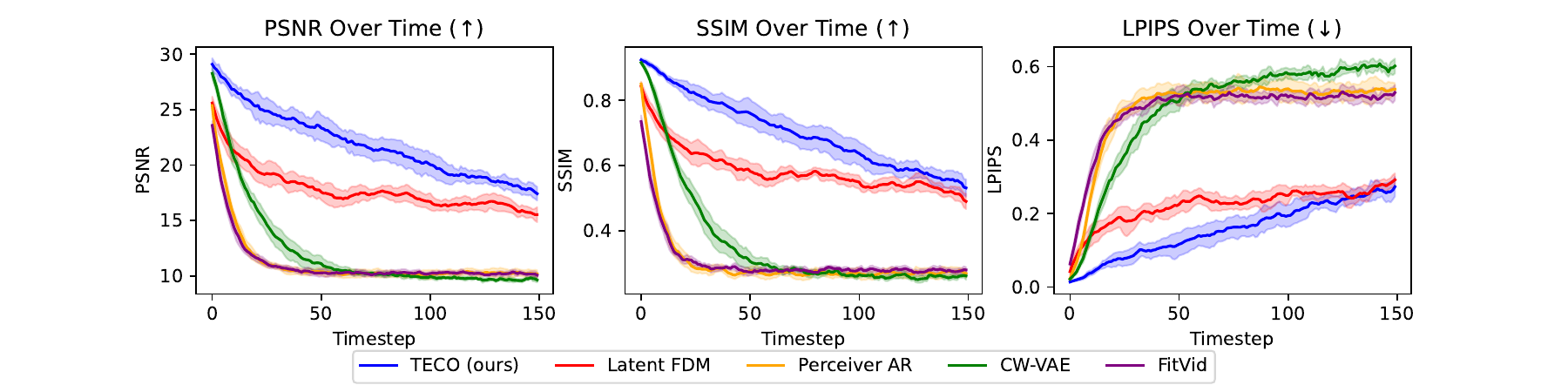}}
\caption{DMLab}
\end{subfigure}\hfill%
\begin{subfigure}[b]{\linewidth}%
\rlap{\includegraphics[width=\linewidth]{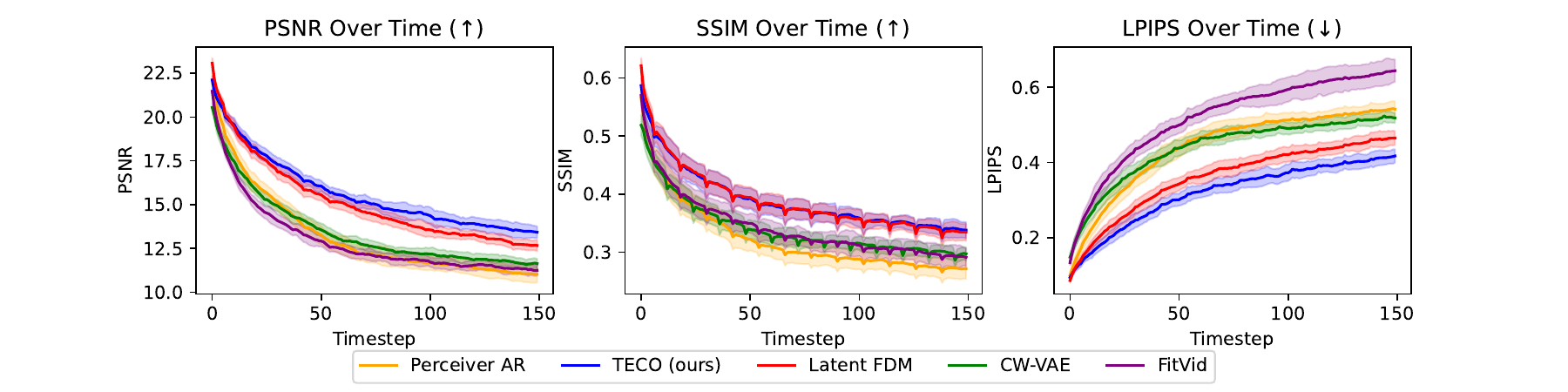}}
\caption{Minecraft}
\end{subfigure}\hfill%
\begin{subfigure}[b]{\linewidth}%
\rlap{\includegraphics[width=\linewidth]{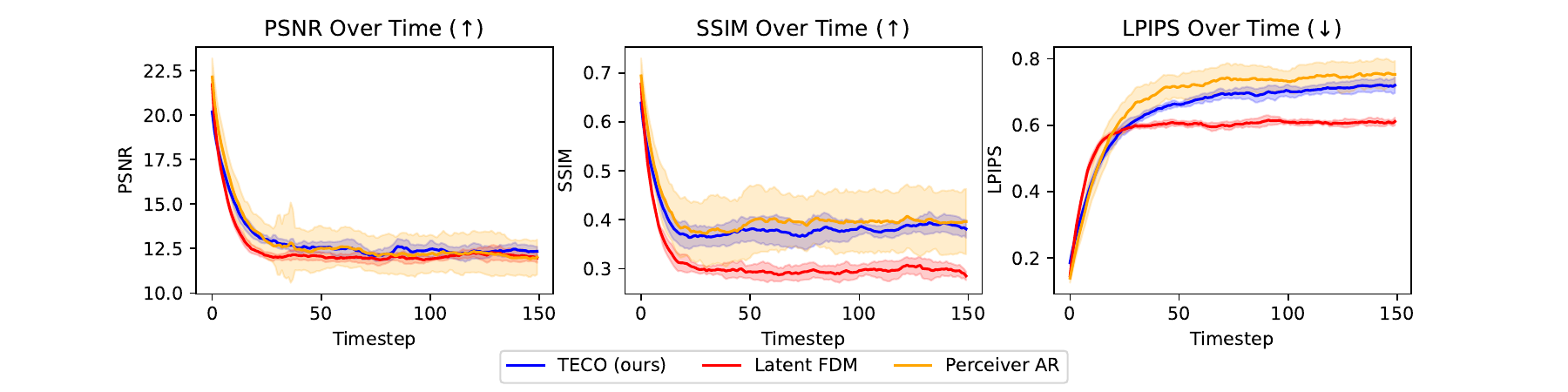}}
\caption{Habitat}
\end{subfigure}\hfill%

\caption{All plots shows PSNR, SSIM, and LPIPS on 150 predicted frames conditioned on 144 frames. The 144 conditioned frames are not shown on the graphs and timestep 0 corresponds to the first predicted frame}
\label{fig:horizon}
\end{figure}
\Cref{fig:horizon} shows PSNR, SSIM, and LPIPS as a function of prediction horizon for each dataset. Generally, each plot reflected the corresponding aggregated metrics in \Cref{table:main}. For DMLab, TECO shows much better temporal consistency for the full trajectory, with Latent FDM coming in second. CW-VAE is able retain some consistency but drops fairly quickly. Lastly, FitVid and Perceiver AR lose consistency very quickly. We see a similar trend in Minecraft, with Latent FDM coming closer in matching TECO. For Habitat, all methods generally have trouble producing consistent predictions, primarily due to the difficulty of the environment.

\clearpage

\section{Performance versus Training Sequence Length}
\label{section:perfseqlength}
\begin{figure}[H]
\begin{subfigure}[b]{0.45\linewidth}
\rlap{\includegraphics[width=\linewidth]{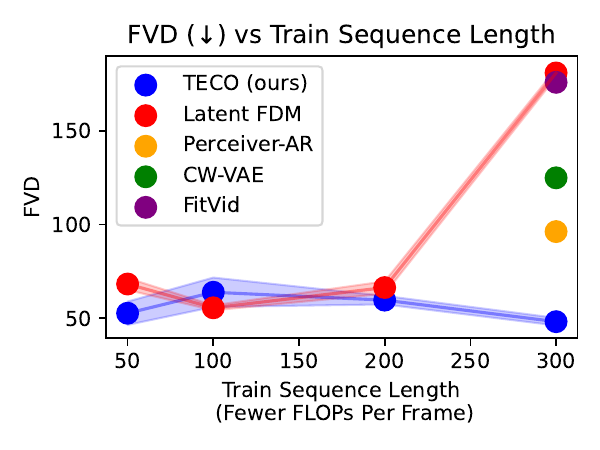}}
\end{subfigure}
\hfill
\begin{subfigure}[b]{0.45\linewidth}
\includegraphics[width=\linewidth]{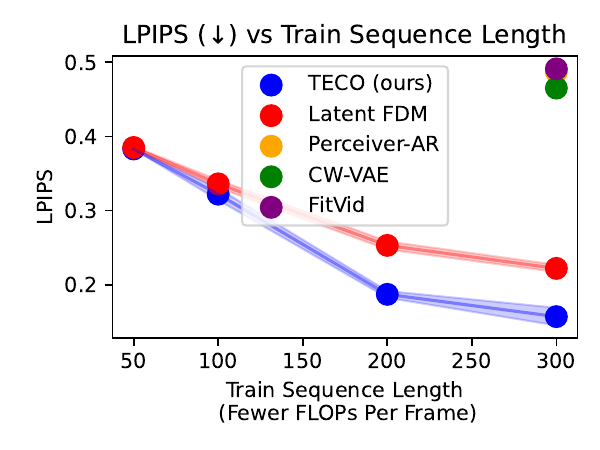}
\end{subfigure}

\begin{subfigure}[b]{0.45\linewidth}
\rlap{\includegraphics[width=\linewidth]{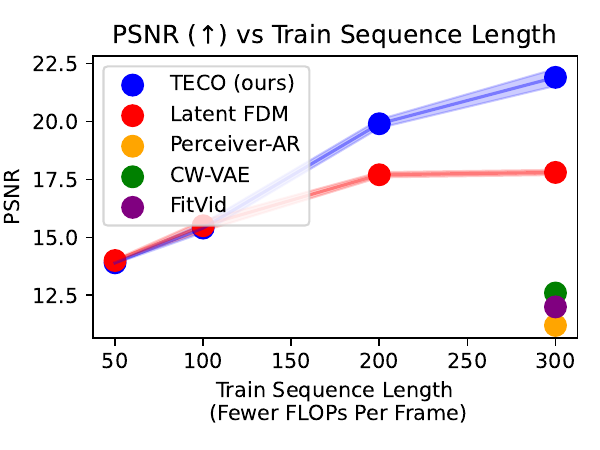}}
\end{subfigure}
\hfill
\begin{subfigure}[b]{0.45\linewidth}
\includegraphics[width=\linewidth]{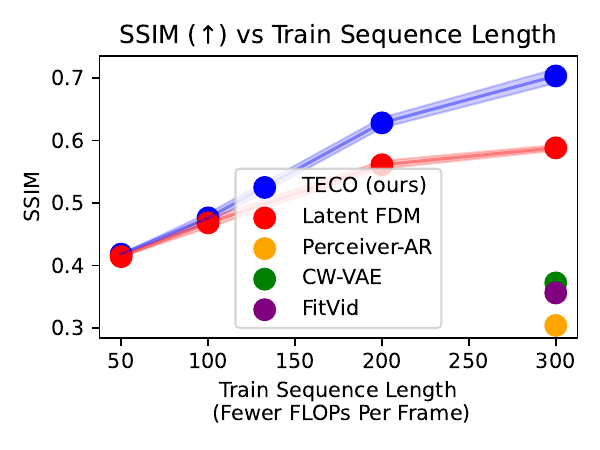}
\end{subfigure}
\caption{DMLab}
\label{fig:scaling_dmlab}
\end{figure}

\clearpage

\begin{figure}[H]
\begin{subfigure}[b]{0.45\linewidth}
\rlap{\includegraphics[width=\linewidth]{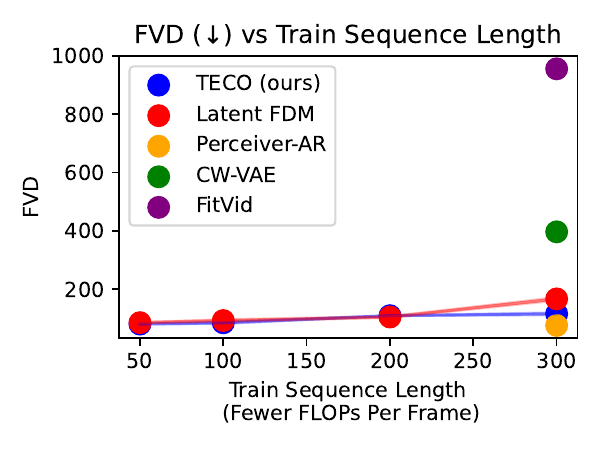}}
\end{subfigure}
\hfill
\begin{subfigure}[b]{0.45\linewidth}
\includegraphics[width=\linewidth]{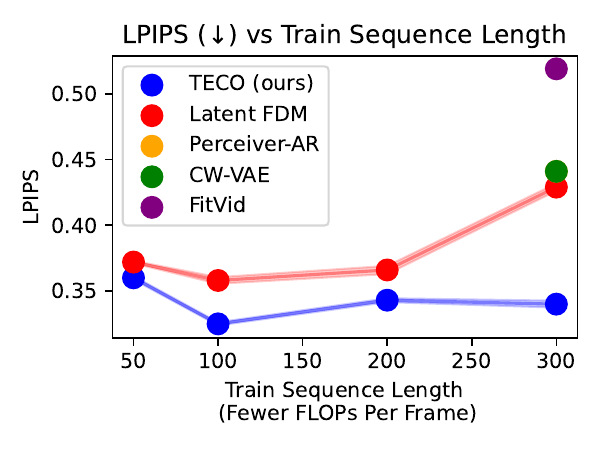}
\end{subfigure}

\begin{subfigure}[b]{0.45\linewidth}
\rlap{\includegraphics[width=\linewidth]{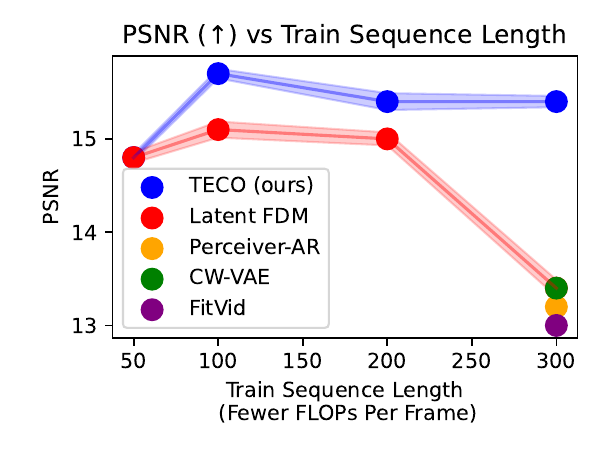}}
\end{subfigure}
\hfill
\begin{subfigure}[b]{0.45\linewidth}
\includegraphics[width=\linewidth]{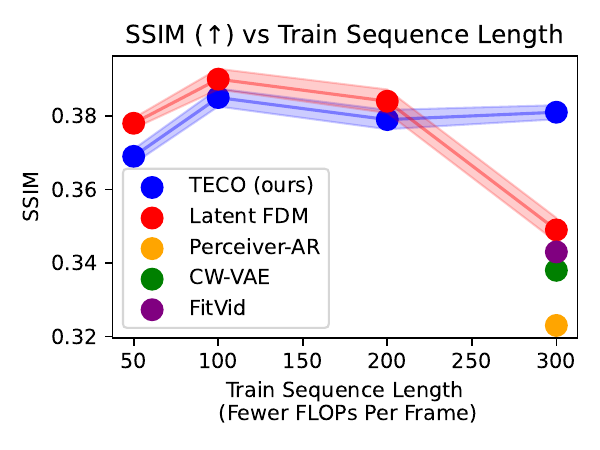}
\end{subfigure}
\caption{Minecraft}
\label{fig:scaling_minecraft}
\end{figure}

\Cref{fig:scaling_dmlab} and \Cref{fig:scaling_minecraft} show plots comparing performance with training models on different sequence lengths. Under a fixed compute budget and batch size, training on shorter videos enables us to scale to larger models. This can also be interpreted as model capacity or FLOPs allocated per image. In general, training on shorter videos enables higher quality frames (per-image) but at a cost of worse temporal consistency due to reduced context length. We can see a very clear trend in DMLab, in that TECO is able to better scale on longer sequences, and correspondingly benefits from it. Latent FDM has trouble when training on full sequences. We hypothesize that this may be due to diffusion models being less amenable towards downsamples, it it needs to model and predict noise. In Minecraft, we see the best performance at around 50-100 training frames, where a model has higher fidelity image predictions, and also has sufficient context.

\clearpage

\section{Sampling Time}
\label{section:samplingtime}
\begin{figure}[H]
\begin{subfigure}[b]{0.5\linewidth}
\rlap{\includegraphics[width=\linewidth]{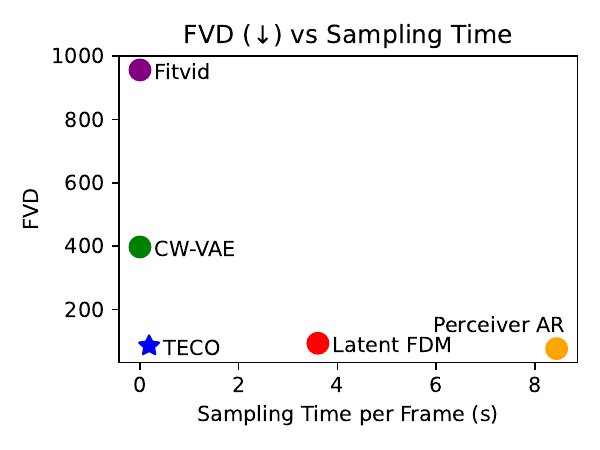}}
\end{subfigure}
\hfill
\begin{subfigure}[b]{0.5\linewidth}
\includegraphics[width=\linewidth]{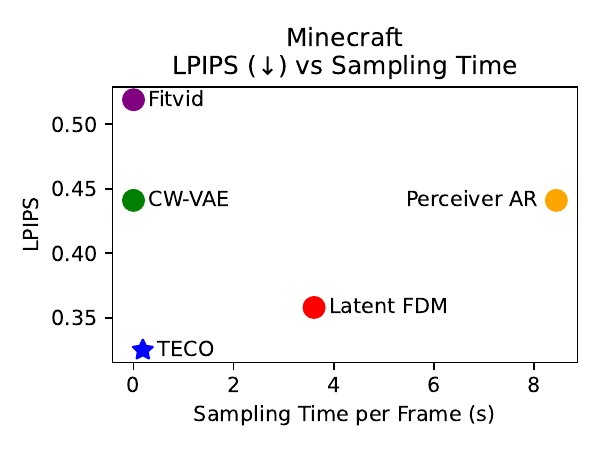}
\end{subfigure}
\end{figure}

\begin{table}[H]
\centering
\begin{tabular}{@{}lc@{}}
\toprule
             & Sampling Time per Frame (ms) \\ \midrule
TECO (ours)  & $186$                        \\
Latent FDM   & $3606$                       \\
Perceiver-AR & $8443$                       \\
CW-VAE       & $0.062$                      \\
FitVid       & $0.074$                      \\ \bottomrule
\end{tabular}
\end{table}

\section{Ablations}
\label{section:ablations}

\begin{table}[H]
    \begin{subtable}[h]{0.6\linewidth}
        \centering
        \begin{tabular}{@{}ccc@{}}
        \toprule
        DropLoss Rate & FVD & Train Step (ms)                    \\ \midrule
        0.8             & $187$ &    $125$                \\
        0.6             & $186$  &  $143$                 \\
        0.4             & $184$  & $155$ \\
        0.2             & $184$   &    $167$              \\
        0.0             & $182$ & $182$ \\ \bottomrule
        \end{tabular}
        \caption{DropLoss Rates}
    \end{subtable}
    \hfill
    \begin{subtable}[h]{0.4\linewidth}
        \centering
        \begin{tabular}{@{}lr@{}}
        \toprule
        Posteriors                                   & FVD                       \\
        \midrule
        VQ (+ MaskGit prior) (ours)                                    & $189$                     \\
        OneHot (+ MaskGit prior) & $199$ \\
        OneHot (+ Block AR prior)                    & $209$ \\
        OneHot (+ Independent prior)                  & $228$                     \\
        Argmax (+ MaskGit prior)                  & $336$                     \\
        \bottomrule
        \end{tabular}
        \caption{Posteriors}
    \end{subtable}
    \begin{subtable}[h]{0.33\linewidth}
        \centering
        \begin{tabular}{@{}lr@{}}
            \toprule
            Dynamics Prior & FVD                       \\ \midrule
            MaskGit (ours) & $189$                     \\
            Independent    & $220$ \\ 
            Autoregressive & $207$                     \\
            \bottomrule
        \end{tabular}
        \caption{Prior Networks}
    \end{subtable}
    \begin{subtable}[h]{0.33\linewidth}
        \centering
        \begin{tabular}{@{}lr@{}}
        \toprule
        Conditional Encoding       & FVD \\ \midrule
        Yes (ours)                    & 189 \\
        No & 208 \\ \bottomrule
        \end{tabular}
        \caption{Conditional Encoding}
    \end{subtable}
    \begin{subtable}[h]{0.33\linewidth}
    \centering
        \begin{tabular}{@{}cc@{}}
            \toprule
            Number of Codes & FVD \\ \midrule
            64              & 191 \\
            256             & 195 \\
            1024            & 186 \\
            4096            & 200 \\ \bottomrule
            \end{tabular}
            \caption{VQ Codebook Size}
    \end{subtable}
    \caption{Ablations comparing alternative prior, posterior, and codebook designs}
    \label{table:abl_arch}
\end{table}

\begin{table}[H]
    \begin{subtable}[h]{0.28\linewidth}
        \centering
        \begin{tabular}{@{}lcl@{}}
        \toprule
                  & \multicolumn{2}{c}{FVD}   \\
                        Size & $2\times 2$ & $4\times 4$ \\ \midrule
        Base            & 204         & 189         \\
        Small Enc & 214         & 191         \\
        Small Dec & 232         & 198         \\ \bottomrule
        \end{tabular}
        \caption{Encoder and Decoder}
    \end{subtable}
    \begin{subtable}[h]{0.36\linewidth}
        \centering
        \begin{tabular}{@{}cccc@{}}
        \toprule
               &       & \multicolumn{2}{c}{FVD}   \\
        Layers & Width & $2\times 2$ & $4\times 4$ \\ \midrule
        8      & 768   & 204         & 189         \\
        8      & 384   & 260         & 196         \\
        2      & 768   & 216         & 202         \\ \bottomrule
        \end{tabular}
        \caption{Temporal Transformer}
    \end{subtable}
    \begin{subtable}[h]{0.36\linewidth}
        \centering
        \begin{tabular}{@{}cccc@{}}
        \toprule
               &       & \multicolumn{2}{c}{FVD}   \\
        Layers & Width & $2\times 2$ & $4\times 4$ \\ \midrule
        8      & 768   & 204         & 189         \\
        8      & 384   & 228         & 193         \\
        2      & 768   & 228         & 201         \\ \bottomrule
        \end{tabular}
        \caption{MaskGit Prior}
    \end{subtable}
    \caption{Ablations on scaling different parts of TECO.}
    \label{table:abl_scaling}
\end{table}

\begin{table}[H]
\centering
\begin{tabular}{@{}lccccc@{}}
\toprule
               & FVD ($\downarrow$) & PSNR ($\uparrow$) & SSIM ($\uparrow$) & LPIPS ($\downarrow$) & Train Step Time (ms) \\ \midrule
TECO (ours)    & $48$             & $\mathbf{21.9}$   & $\mathbf{0.703}$  & $\mathbf{0.157}$     & $\mathbf{151}$       \\
MaskGit        & $950$             & $19.3$            & $0.605$           & $0.274$              & $167$                \\
Autoregressive & $\textbf{44}$    & $20.1$            & $0.640$           & $0.197$              & $267$                \\ \bottomrule
\end{tabular}
\caption{DMLab dataset comparisons against similar model as TECO without latent dynamics, and Maskgit or AR model on VQ-GAN tokens directly.}
\label{table:abl_dynamics}
\end{table}
\Cref{table:abl_dynamics} shows comparisons between TECO and alternative architectures that do not use latent dynamics. Architecturally, MaskGit and Autoregressive are very similar to TECO, with a few small changes: (1) there is no CNN decoder and (2) MaskGit and AR directly predict the VQ-GAN latents (as opposed to the learned VQ latents in TECO). In terms of training time, MaskGit and AR are a little slower since they operate on $16 \times 16$ latents instead of $8\times 8$ latents for TECO. In addition, conditioning for the AR model is done using cross attention, as channel-wise concatenation does not work well due to unidirectioal masking. Both models without latent dynamics have worse temporal consistency, as well as overall sample quality. We hypothesize that TECO has better temporal consistency due to weak bottlenecking of latent representation, as a lot of time can be spent modeling likelihood of imperceptible image / video statistics. MaskGit shows very high FVD due to a tendency to collapse in later frames of prediction, which FVD is sensitive to.

\clearpage

\section{Metrics During Training}
\label{section:metricstraining}
\begin{figure}[H]
    \centering
    \includegraphics[width=\textwidth]{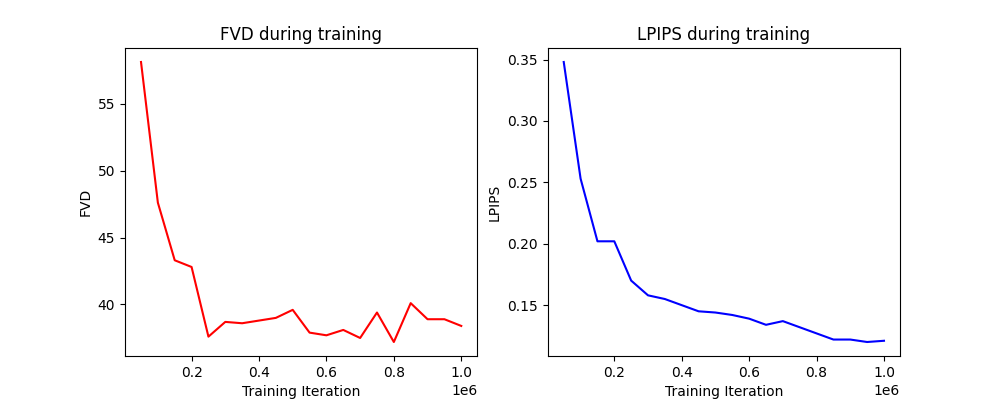}
    \caption{Comparing FVD and LPIPS evaluation metrics over the course of training. FVD tends to saturate earlier (200k) while LPIPS keeps on improving up until 1M iterations.}
    \label{fig:fvd_lpips_ckpts}
\end{figure}
\Cref{fig:fvd_lpips_ckpts} shows plots of FVD (over chunks of generatd 16 frame video) and LPIPS during training, evaluated at saved model checkpoints every 50k iterations over 1M iterations. We can see that although FVD (measuring frame fidelity) tends to saturate early on during training (at around 200k iterations), the long-term consistency metric (LPIPS) continues to improve until the end of training. We hypothesize that this may be due to the model first learning the "easier bits" more local in time, and then learning long-horizon bits once the easier bits have been learned.

\section{High Quality Spatio-Temporal Compression}
\label{section:compression}
\begin{table}[H]
\centering
\begin{tabular}{@{}lll@{}}
\toprule
Model                 & Dataset   & FVD$\downarrow$ \\ \midrule
\multirow{2}{*}{TATS} & DMLab     & 54              \\
                      & Minecraft & 226             \\ \midrule
\multirow{2}{*}{TECO} & DMLab     & \textbf{7}           \\
                      & Minecraft & \textbf{53}          \\ \bottomrule
\end{tabular}
\caption{Reconstruction FVD comparing TATS Video VQGAN to TECO}
\label{table:tats_teco_recon}
\end{table}
\Cref{table:tats_teco_recon} compares reconstruction FVD between TECO and TATS. At the same compression rate (same number of discrete codes), TECO learns far better spatio-temporal codes that TATS, with more of a difference on more visually complex scenes (Minecraft vs DMLab).
\clearpage
\section{Trade-off Between Fidelity and Learning Long-Range Dependencies}
\label{section:tradeoff}
\begin{table}[H]
\centering
\begin{tabular}{@{}lllll@{}}
\toprule
Downsample Resolution & FVD$\downarrow$ & PSNR$\uparrow$ & SSIM$\uparrow$ & LPIPS$\downarrow$ \\ \midrule
$1\times 1$           & 44              & \textbf{20.4}       & \textbf{0.666}      & \textbf{0.170}         \\
$2\times 2$           & 38              & 18.6           & 0.597          & 0.221             \\
$4\times 4$           & \textbf{33}          & 17.7           & 0.578          & 0.242             \\ \bottomrule
\end{tabular}
\caption{Comparing different input resolutions to the temporal transformer}
\label{table:temp_res}
\end{table}

\begin{table}[H]
\centering
\begin{tabular}{@{}lllll@{}}
\toprule
Latent FDM Arch                                   & FVD$\downarrow$ & PSNR$\uparrow$ & SSIM$\uparrow$ & LPIPS$\downarrow$ \\ \midrule
More downsampling + lower resolution computations & 181             & \textbf{17.8}       & \textbf{0.588}      & \textbf{0.222}         \\
Less downsample + higher resolution computations  & \textbf{94}          & 15.6           & 0.501          & 0.277             \\ \bottomrule
\end{tabular}
\caption{omparing different Latent FDM architectures with more computation at different resolutions}
\label{table:latent_fdm_arch}
\end{table}
\Cref{table:temp_res} and \Cref{table:latent_fdm_arch} show a trade-off between fidelity (frame or image quality) and temporal consistency (long-range dependencies) for video prediction architectures (both TECO, and Latent FDM).

\clearpage

\section{Full Experimental Results}
\label{section:full_results}
\begin{table}[H]
\centering
\begin{tabular}{@{}lcccccc@{}}
\toprule
             & TPU-v3 Days & Params & FVD $\downarrow$         & PSNR $\uparrow$           & SSIM $\uparrow$             & LPIPS $\uparrow$            \\ \midrule
TECO (ours)  & $32$        & 169M   & $\mathbf{27.5 \pm 1.77}$ & $\mathbf{22.4 \pm 0.368}$ & $\mathbf{0.709 \pm 0.0119}$ & $\mathbf{0.155 \pm 0.00958}$ \\
Latent FDM   & $32$        & 31M    & $181 \pm 2.20$           & $17.8 \pm 0.111$          & $0.588 \pm 0.00453$         & $0.222 \pm 0.00493$         \\
Perceiver-AR & $32$        & 30M    & $96.3 \pm 3.64$          & $11.2 \pm 0.00381$        & $0.304 \pm 0.0000456$       & $0.487 \pm 0.00123$         \\
CW-VAE       & $32$        & 111M   & $125 \pm 7.95$           & $12.6 \pm 0.0585$         & $0.372 \pm 0.000330$        & $0.465 \pm 0.00156$         \\
FitVid       & $32$        & 165M   & $176 \pm 4.86$           & $12.0 \pm 0.0126$         & $0.356 \pm 0.00171$         & $0.491 \pm 0.00108$         \\ \bottomrule
\end{tabular}
\caption{DMLab}
\end{table}

\begin{table}[H]
\centering
\begin{tabular}{@{}lcccccc@{}}
\toprule
             & TPU-v3 Days & Params & FVD $\downarrow$         & PSNR $\uparrow$            & SSIM $\uparrow$              & LPIPS $\uparrow$             \\ \midrule
TECO (ours)  & $80$        & 274M   & $116\pm 5.08$            & $\mathbf{15.4 \pm 0.0603}$ & $\mathbf{0.381 \pm 0.00192}$ & $\mathbf{0.340 \pm 0.00264}$ \\
Latent FDM   & $80$        & 33M    & $167 \pm 6.26$           & $13.4 \pm 0.0904$          & $0.349 \pm 0.00327$          & $0.429 \pm 0.00284$          \\
Perceiver-AR & $80$        & 166M   & $\mathbf{76.3 \pm 1.72}$ & $13.2 \pm 0.0711$          & $0.323 \pm 0.00336$          & $0.441 \pm 0.00207$          \\
CW-VAE       & $80$        & 140M   & $397 \pm 15.5$           & $13.4 \pm 0.0610$          & $0.338 \pm 0.00274$          & $0.441 \pm 0.00367$          \\
FitVid       & $80$        & 176M   & $956 \pm 15.8$           & $13.0 \pm 0.00895$         & $0.343 \pm 0.00380$          & $0.519 \pm 0.00367$          \\ \bottomrule
\end{tabular}
\caption{Minecraft}
\end{table}

\begin{table}[H]
\centering
\begin{tabular}{@{}lcccccc@{}}
\toprule
             & TPU-v3 Days & Params & FVD $\downarrow$        & PSNR $\uparrow$           & SSIM $\uparrow$              & LPIPS $\uparrow$              \\ \midrule
TECO (ours)  & $275$       & 386M   & $\mathbf{76.3\pm 1.72}$ & $\mathbf{12.8\pm 0.0139}$ & $0.363\pm 0.00122$           & $0.604\pm 0.00451$            \\
Latent FDM   & $275$       & 87M    & $433 \pm 2.67$          & $12.5 \pm 0.0121$         & $0.311 \pm 0.000829$         & $\mathbf{0.582 \pm 0.000492}$ \\
Perceiver-AR & $275$       & 200M   & $164 \pm 12.6$          & $\mathbf{12.8\pm 0.0423}$ & $\mathbf{0.405 \pm 0.00248}$ & $0.676 \pm 0.00282$           \\ \bottomrule
\end{tabular}
\caption{Habitat}
\end{table}

\begin{table}[H]
\centering
\begin{subtable}[h]{0.5\linewidth}
    \begin{tabular}{@{}cccc@{}}
    \toprule
                 & TPU-v3 Days & Params & FVD $\downarrow$        \\ \midrule
    TECO (ours)  & $640$       & 1.09B  & $649 \pm 16.5$          \\
    Latent FDM   & $640$       & 831M   & $960 \pm 52.7$ \\
    Perceiver-AR & $640$       & 1.06B  &  $\mathbf{607 \pm 6.98}$         \\ \bottomrule
    \end{tabular}
    \caption{Using top-k sampling for Perceiver AR and TECO}
\end{subtable}
\centering
\begin{subtable}[h]{0.5\linewidth}
    \begin{tabular}{@{}cccc@{}}
    \toprule
                 & TPU-v3 Days & Params & FVD $\downarrow$        \\ \midrule
    TECO (ours)  & $640$       & 1.09B  & $\mathbf{799 \pm 23.4}$          \\
    Latent FDM   & $640$       & 831M   & $960 \pm 52.7$ \\
    Perceiver-AR & $640$       & 1.06B  & $1022 \pm 32.4$          \\ \bottomrule
    \end{tabular}
    \caption{No top-k sampling}
\end{subtable}
\caption{Kinetics}
\label{table:kinetics_full}
\end{table}

\begin{figure}[H]
    \centering
    \includegraphics[width=0.5\linewidth]{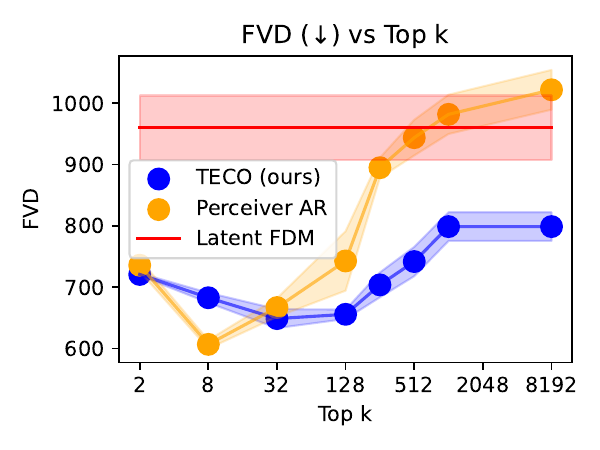}
    \caption{FVD on Kinetics-600 with different top-k values for Perceiver-AR and TECO}
    \label{fig:kinetics_top_k}
\end{figure}
\clearpage

\section{Scaling Results}
\label{section:scaling}

\begin{table}[H]
\centering
\begin{tabular}{@{}cccccccc@{}}
\toprule
                             & \begin{tabular}[c]{@{}c@{}}TPU-v3 \\ Days\end{tabular} & \begin{tabular}[c]{@{}c@{}}Train \\ Seq Len\end{tabular} & Params & FVD $\downarrow$         & PSNR $\uparrow$           & SSIM $\uparrow$             & LPIPS $\downarrow$  \\ \midrule
\multirow{4}{*}{TECO (ours)} & \multirow{4}{*}{$32$}                                  & 300                                                      & 169M   & $\mathbf{48.2 \pm 2.02}$ & $\mathbf{21.9 \pm 0.368}$ & $\mathbf{0.703 \pm 0.0114}$ & $\mathbf{0.157 \pm 0.0119}$  \\
                             &                                                        & 200                                                      & 169M   & $59.7 \pm 2.29$          & $19.9 \pm 0.186$          & $0.628 \pm 0.00821$         & $0.187\pm 0.00460$  \\
                             &                                                        & 100                                                      & 86M    & $63.9\pm 7.84$           & $15.4\pm 0.199$           & $0.476 \pm 0.00745$         & $0.322 \pm 0.00792$ \\
                             &                                                        & 50                                                       & 195M   & $52.7 \pm 6.23$          & $13.9 \pm 0.0311$         & $0.418 \pm 0.000659$        & $0.383\pm 0.000302$ \\ \midrule
\multirow{4}{*}{Latent FDM}  & \multirow{4}{*}{$32$}                                  & 300                                                      & 31M    & $181 \pm 2.20$           & $17.8 \pm 0.111$          & $0.588 \pm 0.00453$         & $0.222 \pm 0.00493$ \\
                             &                                                        & 200                                                      & 62M    & $66.4 \pm 3.31$          & $17.7 \pm 0.114$          & $0.561 \pm 0.00623$         & $0.253 \pm 0.00550$ \\
                             &                                                        & 100                                                      & 80M    & $55.6 \pm 1.36$          & $15.5 \pm 0.233$          & $0.468 \pm 0.00776$         & $0.336 \pm 0.00511$ \\
                             &                                                        & 50                                                       & 110M   & $68.3 \pm 3.19$          & $14.0 \pm 0.0445$         & $0.414 \pm 0.424$           & $0.385 \pm 0.00151$ \\ \bottomrule
\end{tabular}
\caption{DM Lab scaling}
\end{table}

\begin{table}[H]
\centering
\begin{tabular}{@{}cccccccc@{}}
\toprule
                             & \begin{tabular}[c]{@{}c@{}}TPU-v3 \\ Days\end{tabular} & \begin{tabular}[c]{@{}c@{}}Train \\ Seq Len\end{tabular} & Params & FVD $\downarrow$ & PSNR $\uparrow$            & SSIM $\uparrow$              & LPIPS $\downarrow$           \\ \midrule
\multirow{4}{*}{TECO (ours)} & \multirow{4}{*}{$80$}                                  & 300                                                      & 274M   & $116 \pm 5.08$   & $15.4 \pm 0.0603$          & $0.381 \pm 0.00192$          & $0.340 \pm 0.00264$          \\
                             &                                                        & 200                                                      & 261M   & $109.5 \pm 1.46$ & $15.4 \pm 0.0906$          & $0.379 \pm 0.00263$          & $0.343 \pm 0.00148$          \\
                             &                                                        & 100                                                      & 257M   & $85.1 \pm 4.09$  & $\mathbf{15.7 \pm 0.0516}$ & $0.385 \pm 0.00244$          & $\mathbf{0.325 \pm 0.00121}$ \\
                             &                                                        & 50                                                       & 140M   & $80.7 \pm 1.42$  & $14.8 \pm 0.0404$          & $0.369 \pm 0.00197$          & $0.360 \pm 0.00133$          \\ \midrule
\multirow{4}{*}{Latent FDM}  & \multirow{4}{*}{$80$}                                  & 300                                                      & 33M    & $167 \pm 6.26$   & $13.4 \pm 0.0904$          & $0.349 \pm 0.00327$          & $0.429 \pm 0.00284$          \\
                             &                                                        & 200                                                      & 80M    & $104.9 \pm 3.21$ & $15.0 \pm 0.0701$          & $0.384 \pm 0.00320$          & $0.366 \pm 0.00311$          \\
                             &                                                        & 100                                                      & 69M    & $92.8 \pm 4.40$  & $15.1 \pm 0.0866$          & $\mathbf{0.390 \pm 0.00281}$ & $0.358 \pm 0.00250$          \\
                             &                                                        & 50                                                       & 186M   & $85.6 \pm 2.25$  & $14.8 \pm 0.0578$          & $0.378 \pm 0.00144$          & $0.372 \pm 0.000966$         \\ \bottomrule
\end{tabular}
\caption{Minecraft scaling}
\end{table}
\clearpage

\section{Related Work}
\label{section:related_work}
\textbf{Video Generation}\quad
Prior video generation methods can be divided into a few classes of models: variational models, exact likelihood models, and GANs. SV2P~\citep{babaeizadeh2017stochastic}, SVP~\citep{denton2018stochastic}, SVG~\citep{villegas2019high}, and FitVid~\cite{babaeizadeh2021fitvid} are variational video generation methods models videos through stochastic latent dynamics, optimized using the ELBO~\citep{kingma2013auto} objective extended in time. SAVP~\citep{lee2018stochastic} adds an adversarial~\citep{goodfellow2014generative} loss to encourage more realistic and high-fidelity generation quality. Diffusion models~\citep{ho2020denoising,sohldickstein2014thermodynamics} have recently emerged as a powerful class of variational generative models which learn to iteratively denoise an initial noise sample to generate high-quality images. There have been several recent works that extend diffusion models to video, through temporal attention~\citep{ho2022video,harvey2022flexible}, 3D convolutions~\citep{hoppe2022diffusion}, or channel stacking~\citep{voleti2022masked}. Unlike variational models, autoregressive models (AR) and flows~\citep{kumar2019videoflow} model videos by optimizing exact likelihood. Video Pixel Networks~\citep{kalchbrenner2017video} and Subscale Video Transformers~\citep{weissenborn2019scaling} autoregressively model each pixel. For more compute efficient training, some prior methods~\citep{yan2021videogpt,le2021ccvs,seo2022harp,rakhimov2020latent,walker2021predicting} propose to learn an AR model in a spatio-temporally compressed latent space of a discrete autoencoder, which has shown to be orders of magnitudes more efficient compared to pixel-based methods. Instead of a VQ-GAN, \cite{le2021ccvs}, learns a frame conditional autoencoder through a flow mechanism. Lastly, GANs~\citep{goodfellow2014generative} offer an alternative method to training video models. MoCoGAN~\citep{tulyakov2018mocogan} generates videos by disentangling style and motion. MoCoGAN-HD~\citep{tian2021good} can efficiently extend to larger resolutions by learning to navigate the latent space of a pretrained image generator. TGANv2~\citep{saito2018tganv2}, DVD-GAN~\citep{clark2019adversarial}, StyleGAN-V~\citep{skorokhodov2021stylegan}, and TrIVD-GAN~\citep{luc2020transformation} introduce various methods to scale to complex video, such as proposing sparse training, or more efficient discriminator design. 

The main focus of this work lies with video prediction, a specific interpretation of conditional video generation. Most prior methods are trained autoregressive in time, so they can be easily extended to video prediction. Video Diffusion, although trained unconditionally proposes reconstruction guidance for prediction. GANs generally require training a separate model for video prediction. However, some methods such as MoCoGAN-HD  and DI-GAN can approximate frame conditioning by inverting the generator to compute a corresponding latent for a frame.

\textbf{Long-Horizon Video Generation}\quad
CW-VAE~\citep{saxena2021clockwork} learns a hierarchy of stochastic latents to better model long term temporal dynamics, and is able to generate videos with long-term consistency for hundreds of frames. TATS~\citep{ge2022long} extends VideoGPT which allows for sampling of arbitrarily long videos using a sliding window. In addition, TATs and CogVideo~\citep{hong2022cogvideo} propose strided sampling as a simple method to incorporate longer horizon contexts. StyleGAN-V~\citep{skorokhodov2021stylegan} and DI-GAN~\citep{yu2022generating} learn continuous-time representations for videos which allow for sampling of arbitrary long videos as well. \cite{brooks2022generating} proposes an efficient video GAN architecture that is able to generate high resolution videos of 128 frames on complex video data for dynamic scenes and horseback riding. FDM~\citep{harvey2022flexible} proposes a diffusion model that is trained to be able to flexibly condition on a wide range of sampled frames to better incorporate context of arbitrarily long videos. \cite{lee2021revisiting} is able to leverage a hierarchical prediction framework using semantic segmentations to generate long videos.

\textbf{Long-Horizon Video Understanding}\quad
Outside of generative modeling, prior work such as MeMViT~\citep{wu2022memvit} and Vis4mer~\citep{mohaiminul2022long} introduce architectures for modeling long-horizon dependencies in videos.

\clearpage

\section{Dataset Details}
\label{section:dataset_details}
\subsection{DMLab}
\label{section:dataset_dmlab_details}
We generate random $7\times 7$ mazes split into four quadrants, with each quadrant containing a random combination of wall and floor textures. We generate 40k trajectories of 300 frames, each $64\times 64$ images. Actions in this environment consist of $20^{\circ}$ left turn, $20^{\circ}$ right turn, and walk forward. In order to maximally traverse the maze, we code an agent that traverses to the furthest unvisited point in the maze, with some added noise for stochasticity. Since the maze is a grid, we can easily hard-code a navigation policy to move to any specified point in the maze.

For 3D visualizations, we also collect depth, camera intrinsics and camera extrinsics (pose) for each timestep. Given this information, we can project RGB points into a 3D coordinate space and reconstruct the maze as a 3D pointcloud. Note that since videos are generated \text{only using RGB} as input, they do not have groundtruth depth and pose. Therefore, we train depth and pose estimators that are used during evaluation. Specifically, we train a depth estimator to map from RGB frame to depth, and a pose estimator that takes in two adjacent RGB frames and predicts the relative change in orientation. During evaluation, we are given an initial ground truth orientation that we apply sequentially to predicted frames.

Although the GQN Mazes~\citep{eslami2018neural} already exists as a video prediction dataset, it is difficult to properly measure temporal consistency. The 3D scenes are relatively simple, and it does not have actions to help reduce stochasticity in using metrics such as PSNR, SSIM, and LPIPS. As a result, FVD is the reliable metric used in GQN Mazes, but tends to be sensitive to noise in video predictions. In addition, we perform 3D visualizations using our dataset that are not possible with GQN Mazes.

\subsection{Minecraft}
We generate 200k trajectories (each of a different Minecraft world) of 300 $128\times 128$ frames in the Minecraft marsh biome. We hardcode an agent to randomly traverse the surroundings by taking left, right, and forward actions with different probabilities. In addition, we let the agent constantly jump, which we found to help traverse simple hills, and prevent itself from drowning. We specifically chose the marsh biome, as it contains hilly turns with sparse collections of trees that act as clear landmarks for consistent generation. Forest and jungle biomes tend to be too dense for any meanginfully clear consistency, as all surroundings look nearly identical. On the other hand, plains biomes had the opposite issue where the surroundings were completely flat. Mountain biomes were too hilly and difficult to traverse.

We opt to introduce an alternative to the MineRL Navigate~\citep{guss2019minerl} since this dataset primarily consists of human demonstrations of people navigating to specific points. This means that trajectories usually follow a relatively straight line, so there are not many long-term dependencies in this dataset, as only a few past frames of context are necessary for prediction.

\subsection{Habitat}
Habitat is a 3D simulator that can render realistic trajectories in scans of 3D scenes. We compile roughly 1400 3D scans from HM3D~\citep{ramakrishnan2021habitat}, MatterPort3D~\citep{chang2017matterport3d} and Gibson~\citep{xia2018gibson}, and generate a total of 200k trajectories of $300$ $128 \times 128$ frames. We use the in-built path traversal algorithm provided in Habitat to construct action trajectories that allow our agent to move between randomly sampled locations in the 3D scene. Similar to Minecraft and DMLab, the agent action space consists of left turn, right turn, and move forward.

\clearpage

\section{Hyperparameters}
\label{section:hyperparams}
\subsection{VQ-GAN \& VAE}
\begin{table}[H]
\centering
\begin{tabular}{@{}lcc@{}}
\toprule
                       & DMLab / Minecraft       & Habitat / Kinetics-600  \\ \midrule
GPU Days               & 16                      & 32                      \\
Resolution             & 64 / 128                & 128                     \\
Batch Size             & 64                      & 64                      \\
LR                     & $3\times 10^{-4}$       & $3\times 10^{-4}$       \\
Num Res Blocks           & 2                       & 2                       \\
Attention Resolutions  & 16                      & 16                      \\
Channel Mult           & 1,2,2,2                 & 1,2,3,4                 \\
Base Channels          & 128                     & 128                     \\
Latent Size (VQ-GAN)   & $16 \times 16$          & $16 \times 16$          \\
Embedding Dim (VQ-GAN) & 256                     & 256                     \\
Codebook Size (VQ-GAN) & 1024                    & 8192                    \\
Latent Size (VAE)      & $16 \times 16 \times 4$ & $16 \times 16 \times 8$ \\ \bottomrule
\end{tabular}
\end{table}

\subsection{TECO}
\begin{table}[H]
\centering
\begin{tabular}{@{}llcccc@{}}
\toprule
\multicolumn{2}{c}{Hyperparameters}                                                                    & DMLab             & Minecraft         & Habitat           & Kinetics-600      \\ \midrule
                                                                                & TPU-v3 Days          & 32                & 80                & 275               & 640               \\
                                                                                & Params               & 169M              & 274M              & 386M              & 1.09B             \\
                                                                                & Resolution           & 64                & 128               & 128               & 128               \\
                                                                                & Batch Size           & 32                & 32                & 32                & 32                \\
                                                                                & Sequence Length      & 300               & 300               & 300               & 100               \\
                                                                                & LR                   & $1\times 10^{-4}$ & $1\times 10^{-4}$ & $1\times 10^{-4}$ & $1\times 10^{-4}$ \\
                                                                                & LR Schedule          & cosine            & cosine            & cosine            & cosine            \\
                                                                                & Warmup Steps         & 10k               & 10k               & 10k               & 10k               \\
                                                                                & Total Training Steps & 1M                & 1M                & 1M                & 1M                \\
                                                                                & DropLoss Rate        & 0.9               & 0.9               & 0.9               & 0.9               \\ \midrule
Encoder                                                                         & Depths               & 256, 512          & 256, 512          & 256, 512          & 256, 512          \\
                                                                                & Blocks               & 2                 & 4                 & 4                 & 8                 \\ \midrule
\multirow{2}{*}{Codebook}                                                       & Size                 & 1024              & 1024              & 1024              & 1024              \\
                                                                                & Embedding Dim        & 32                & 32                & 32                & 32                \\ \midrule
Decoder                                                                         & Depths               & 256, 512          & 256, 512          & 256, 512          & 256, 512          \\
                                                                                & Blocks               & 4                 & 8                 & 8                 & 10                \\ \midrule
\multirow{6}{*}{\begin{tabular}[c]{@{}l@{}}Temporal\\ Transformer\end{tabular}} & Downsample Factor    & 8                 & 8                 & 4                 & 2                 \\
                                                                                & Hidden Dim           & 1024              & 1024              & 1024              & 1536              \\
                                                                                & Feedforward Dim      & 4096              & 4096              & 4096              & 6144              \\
                                                                                & Heads                & 16                & 16                & 16                & 24                \\
                                                                                & Layers               & 8                 & 12                & 8                 & 24                \\
                                                                                & Dropout              & 0                 & 0                 & 0                 & 0                 \\ \midrule
\multirow{6}{*}{MaskGit}                                                        & Mask Schedule        & cosine            & cosine            & cosine            & cosine            \\
                                                                                & Hidden Dim           & 512               & 768               & 1024              & 1024              \\
                                                                                & Feedforward Dim      & 2048              & 3072              & 4096              & 4096              \\
                                                                                & Heads                & 8                 & 12                & 16                & 16                \\
                                                                                & Layers               & 8                 & 6                 & 16                & 24                \\
                                                                                & Dropout              & 0                 & 0                 & 0                 & 0                 \\ \bottomrule
\end{tabular}
\end{table}

\begin{table}[H]
\centering
\begin{tabular}{@{}llcccc@{}}
\toprule
                                                                                &                      & \multicolumn{4}{c}{\begin{tabular}[c]{@{}c@{}}Train Sequence Length\\ (Fewer FLOPs per Frame)\end{tabular}} \\
\multicolumn{2}{c}{Hyperparameters}                                                                    & 300                       & 200                       & 100                      & 50                       \\ \midrule
                                                                                & TPU-v3 Days          & 32                        & 32                        & 32                       & 32                       \\
                                                                                & Params               & 169M                      & 169M                      & 86M                      & 195M                     \\
                                                                                & Resolution           & 64                        & 64                        & 64                       & 64                       \\
                                                                                & Batch Size           & 32                        & 32                        & 32                       & 32                       \\
                                                                                & LR                   & $1\times 10^{-4}$         & $1\times 10^{-4}$         & $1\times 10^{-4}$        & $1\times 10^{-4}$        \\
                                                                                & LR Schedule          & cosine                    & cosine                    & cosine                   & cosine                   \\
                                                                                & Warmup Steps         & 10k                       & 10k                       & 10k                      & 10k                      \\
                                                                                & Total Training Steps & 1M                        & 1M                        & 1M                       & 1M                       \\
                                                                                & DropLoss Rate        & 0.9                       & 0.85                      & 0.85                     & 0.85                     \\ \midrule
Encoder                                                                         & Depths               & 256, 512                  & 256, 512                  & 256, 512                 & 256, 512                 \\
                                                                                & Blocks               & 2                         & 2                         & 2                        & 2                        \\ \midrule
\multirow{2}{*}{Codebook}                                                       & Size                 & 1024                      & 1024                      & 1024                     & 1024                     \\
                                                                                & Embedding Dim        & 32                        & 32                        & 32                       & 32                       \\ \midrule
Decoder                                                                         & Depths               & 256, 512                  & 256, 512                  & 256, 512                 & 256, 512                 \\
                                                                                & Blocks               & 4                         & 4                         & 4                        & 4                        \\ \midrule
\multirow{6}{*}{\begin{tabular}[c]{@{}l@{}}Temporal\\ Transformer\end{tabular}} & Downsample Factor    & 8                         & 8                         & 2                        & 2                        \\
                                                                                & Hidden Dim           & 1024                      & 1024                      & 512                      & 1024                     \\
                                                                                & Feedforward Dim      & 4096                      & 4096                      & 2048                     & 4096                     \\
                                                                                & Heads                & 16                        & 16                        & 8                        & 16                       \\
                                                                                & Layers               & 8                         & 8                         & 8                        & 8                        \\
                                                                                & Dropout              & 0                         & 0                         & 0                        & 0                        \\ \midrule
\multirow{6}{*}{MaskGit}                                                        & Mask Schedule        & cosine                    & cosine                    & cosine                   & cosine                   \\
                                                                                & Hidden Dim           & 512                       & 512                       & 512                      & 768                      \\
                                                                                & Feedforward Dim      & 2048                      & 2048                      & 2048                     & 3072                     \\
                                                                                & Heads                & 8                         & 8                         & 8                        & 12                       \\
                                                                                & Layers               & 8                         & 8                         & 8                        & 8                        \\
                                                                                & Dropout              & 0                         & 0                         & 0                        & 0                        \\ \bottomrule
\end{tabular}
\caption{Hyperparameters for scaling TECO on DMLab}
\end{table}

\begin{table}[H]
\centering
\begin{tabular}{@{}llcccc@{}}
\toprule
                                                                                &                      & \multicolumn{4}{c}{\begin{tabular}[c]{@{}c@{}}Train Sequence Length\\ (Fewer FLOPs per Frame)\end{tabular}} \\
Hyperparameters &                                                                    & 300                       & 200                       & 100                      & 50                       \\ \midrule
                                                                                & TPU-v3 Days          & 80                        & 80                        & 80                       & 80                       \\
                                                                                & Params               & 274M                      & 261M                      & 257M                     & 140M                     \\
                                                                                & Resolution           & 128                       & 128                       & 128                      & 128                      \\
                                                                                & Batch Size           & 32                        & 32                        & 32                       & 32                       \\
                                                                                & LR                   & $1\times 10^{-4}$         & $1\times 10^{-4}$         & $1\times 10^{-4}$        & $1\times 10^{-4}$        \\
                                                                                & LR Schedule          & cosine                    & cosine                    & cosine                   & cosine                   \\
                                                                                & Warmup Steps         & 10k                       & 10k                       & 10k                      & 10k                      \\
                                                                                & Total Training Steps & 1M                        & 1M                        & 1M                       & 1M                       \\
                                                                                & DropLoss Rate        & 0.9                       & 0.85                      & 0.25                     & 0.25                     \\ \midrule
Encoder                                                                         & Depths               & 256, 512                  & 256, 512                  & 256, 512                 & 256, 512                 \\
                                                                                & Blocks               & 4                         & 4                         & 4                        & 4                        \\ \midrule
\multirow{2}{*}{Codebook}                                                       & Size                 & 1024                      & 1024                      & 1024                     & 1024                     \\
                                                                                & Embedding Dim        & 32                        & 32                        & 32                       & 32                       \\ \midrule
Decoder                                                                         & Depths               & 256, 512                  & 256, 512                  & 256, 512                 & 256, 512                 \\
                                                                                & Blocks               & 8                         & 8                         & 8                        & 8                        \\ \midrule
\multirow{6}{*}{\begin{tabular}[c]{@{}l@{}}Temporal\\ Transformer\end{tabular}} & Downsample Factor    & 8                         & 4                         & 2                        & 2                        \\
                                                                                & Hidden Dim           & 1024                      & 1024                      & 1024                     & 512                      \\
                                                                                & Feedforward Dim      & 4096                      & 4096                      & 4096                     & 2048                     \\
                                                                                & Heads                & 16                        & 16                        & 16                       & 8                        \\
                                                                                & Layers               & 12                        & 12                        & 12                       & 12                       \\
                                                                                & Dropout              & 0                         & 0                         & 0                        & 0                        \\ \midrule
\multirow{6}{*}{MaskGit}                                                        & Mask Schedule        & cosine                    & cosine                    & cosine                   & cosine                   \\
                                                                                & Hidden Dim           & 768                       & 768                       & 768                      & 768                      \\
                                                                                & Feedforward Dim      & 3072                      & 3072                      & 3072                     & 3072                     \\
                                                                                & Heads                & 12                        & 12                        & 12                       & 12                       \\
                                                                                & Layers               & 6                         & 6                         & 6                        & 8                        \\
                                                                                & Dropout              & 0                         & 0                         & 0                        & 0                        \\ \bottomrule
\end{tabular}
\caption{Hyperparameters for scaling TECO on Minecraft}
\end{table}

\subsection{Latent FDM}

\begin{table}[H]
\centering
\begin{tabular}{@{}lcccc@{}}
\toprule
Hyperparameters       & DMLab             & Minecraft         & Habitat           & Kinetics-600      \\ \midrule
TPU-v3 Days           & 32                & 80                & 275               & 640               \\
Params                & 31M               & 33M               & 87M               & 831M              \\
Resolution            & 64                & 128               & 128               & 128               \\
Batch Size            & 32                & 32                & 32                & 32                \\
LR                    & $1\times 10^{-4}$ & $1\times 10^{-4}$ & $1\times 10^{-4}$ & $1\times 10^{-4}$ \\
LR Schedule           & cosine            & cosine            & cosine            & cosine            \\
Optimizer             & Adam              & Adam              & Adam              & Adam              \\
Warmup Steps          & 10k               & 10k               & 10k               & 10k               \\
Total Training Steps  & 1M                & 1M                & 1M                & 1M                \\
Base Channels         & 128               & 128               & 128               & 256               \\
Num Res Blocks        & 1,1,1,2           & 1,1,2,2           & 1,2,2,4           & 2,2,2,2           \\
Head Dim              & 64                & 64                & 64                & 64                \\
Attention Resolutions & 4,2               & 4,2               & 4,2               & 8,4,2             \\
Dropout               & 0                 & 0                 & 0                 & 0                 \\
Channel Mult          & 1,1,1,2           & 1,2,2,2           & 1,2,2,4           & 1,2,3,8           \\ \bottomrule
\end{tabular}
\caption{Hyperparameters for Latent FDM}
\end{table}

\begin{table}[H]
\centering
\begin{tabular}{@{}lcccc@{}}
\toprule
                      & \multicolumn{4}{c}{\begin{tabular}[c]{@{}c@{}}Train Sequence Length\\ (Fewer FLOPs per Frame)\end{tabular}} \\
Hyperparameters       & 300                       & 200                       & 100                      & 50                       \\ \midrule
TPU-v3 Days           & 32                        & 32                        & 32                       & 32                       \\
Params                & 31M                       & 62M                       & 80M                      & 110M                     \\
Resolution            & 64                        & 64                        & 64                       & 64                       \\
Batch Size            & 32                        & 32                        & 32                       & 32                       \\
LR                    & $1\times 10^{-4}$         & $1\times 10^{-4}$         & $1\times 10^{-4}$        & $1\times 10^{-4}$        \\
LR Schedule           & cosine                    & cosine                    & cosine                   & cosine                   \\
Optimizer             & Adam                      & Adam                      & Adam                     & Adam                     \\
Warmup Steps          & 10k                       & 10k                       & 10k                      & 10k                      \\
Total Training Steps  & 1M                        & 1M                        & 1M                       & 1M                       \\
Base Channels         & 128                       & 128                       & 128                      & 192                      \\
Num Res Blocks        & 1,1,1,2                   & 1,1,2,2,4                 & 2,2,2,2                  & 3,3,3,3                  \\
Head Dim              & 64                        & 64                        & 64                       & 64                       \\
Attention Resolutions & 4,2                       & 4,1                       & 4,2                      & 8,4,2                    \\
Dropout               & 0                         & 0                         & 0                        & 0                        \\
Channel Mult          & 1,1,1,2                   & 1,1,2,2,4                 & 1,2,3,4                  & 1,2,3,4                  \\ \bottomrule
\end{tabular}
\caption{Hyperparameters for scaling Latent FDM on DMLab}
\end{table}

\begin{table}[H]
\centering
\begin{tabular}{@{}lcccc@{}}
\toprule
                      & \multicolumn{4}{c}{\begin{tabular}[c]{@{}c@{}}Train Sequence Length\\ (Fewer FLOPs per Frame)\end{tabular}} \\
Hyperparameters       & 300                       & 200                       & 100                      & 50                       \\ \midrule
TPU-v3 Days           & 80                        & 80                        & 80                       & 80                       \\
Params                & 33M                       & 80M                       & 69M                      & 186M                     \\
Resolution            & 128                       & 128                       & 128                      & 128                      \\
Batch Size            & 32                        & 32                        & 32                       & 32                       \\
LR                    & $1\times 10^{-4}$         & $1\times 10^{-4}$         & $1\times 10^{-4}$        & $1\times 10^{-4}$        \\
LR Schedule           & cosine                    & cosine                    & cosine                   & cosine                   \\
Optimizer             & Adam                      & Adam                      & Adam                     & Adam                     \\
Warmup Steps          & 10k                       & 10k                       & 10k                      & 10k                      \\
Total Training Steps  & 1M                        & 1M                        & 1M                       & 1M                       \\
Base Channels         & 128                       & 128                       & 128                      & 192                      \\
Num Res Blocks        & 1,1,2,2                   & 2,2,2,2                   & 3,3,3,3                  & 2,2,2,2                  \\
Head Dim              & 64                        & 64                        & 64                       & 64                       \\
Attention Resolutions & 4,2                       & 4,2                       & 8,4,2                    & 8,4,2                    \\
Dropout               & 0                         & 0                         & 0                        & 0                        \\
Channel Mult          & 1,2,2,2                   & 1,2,3,4                   & 1,2,2,3                  & 1,2,3,4                  \\ \bottomrule
\end{tabular}
\caption{Hyperparameters for scaling Latent FDM on Minecraft}
\end{table}
\clearpage

\subsection{CW-VAE}
\begin{table}[H]
\centering
\begin{tabular}{@{}llcc@{}}
\toprule
\multicolumn{2}{l}{Hyperparameters}              & DMLab             & Minecraft         \\ \midrule
                          & TPU-v3 Days          & 32                & 80                \\
                          & Params               & 111M              & 140M              \\
                          & Resolution           & 64                & 128               \\
                          & Batch Size           & 32                & 32                \\
                          & LR                   & $1\times 10^{-4}$ & $1\times 10^{-4}$ \\
                          & LR Schedule          & cosine            & cosine            \\
                          & Optimizer            & Adam              & Adam              \\
                          & Warmup Steps         & 10k               & 10k               \\
                          & Total Training Steps & 1M                & 1M                \\ \midrule
\multirow{2}{*}{Encoder}  & Kernels              & 4,4,4             & 4,4,4             \\
                          & Filters              & 256,512,1024      & 256,512,1024      \\ \midrule
\multirow{2}{*}{Decoder}  & Depths               & 256,512           & 256,512           \\
                          & Blocks               & 4                 & 8                 \\ \midrule
\multirow{8}{*}{Dynamics} & Levels               & 3                 & 3                 \\
                          & Abs Factor           & 6                 & 6                 \\
                          & Enc Dense Layers     & 3                 & 3                 \\
                          & Enc Dense Embed      & 1024              & 1024              \\
                          & Cell Stoch Size      & 128               & 256               \\
                          & Cell Deter Size      & 1024              & 1024              \\
                          & Cell Embed Size      & 1024              & 1024              \\
                          & Cell Min Stddev      & 0.001             & 0.001             \\ \bottomrule
\end{tabular}
\caption{Hyperparameters for CW-VAE}
\end{table}

\subsection{FitVid}
\begin{table}[H]
\centering
\begin{tabular}{@{}lcc@{}}
\toprule
Hyperparameters      & DMLab             & Minecraft         \\ \midrule
TPU-v3 Days          & 32                & 80                \\
Params               & 165M              & 176M              \\
Resolution           & 64                & 128               \\
Batch Size           & 32                & 32                \\
LR                   & $1\times 10^{-4}$ & $1\times 10^{-4}$ \\
LR Schedule          & cosine            & cosine            \\
Optimizer            & Adam              & Adam              \\
Warmup Steps         & 10k               & 10k               \\
Total Training Steps & 1M                & 1M                \\
g Dim                & 256               & 256               \\
RNN Size             & 512               & 768               \\
z Dim                & 64                & 128               \\
Filters              & 128,128,256,512   & 128,128,256,512   \\ \bottomrule
\end{tabular}
\caption{Hyperparameters for FitVid}
\end{table}

\end{document}